\documentclass[10pt,twocolumn,letterpaper]{article}

\usepackage[pagenumbers]{cvpr} %

\usepackage{graphicx}
\usepackage{adjustbox}
\usepackage{ragged2e}
\usepackage{tabularx}
\usepackage[most]{tcolorbox}
\usepackage{soul}
\sethlcolor{blue!10!white}
\usepackage{makecell}
\usepackage{ftnright}

\usepackage{booktabs}

\newtcolorbox{formattedquote}{
    colback=blue!3!white,
    colframe=blue!20!white,
    fontupper=\ttfamily\footnotesize,
    boxsep=-5pt %
}

\newtcolorbox{formattedresponse}{colback=yellow!3!white, colframe=yellow!50!white, fontupper=\ttfamily\small}

\definecolor{light}{rgb}{0.5, 0.5, 0.5}
\def\light#1{{\color{light}#1}}
\definecolor{fushia}{rgb}{102, 0, 204}

\definecolor{forestgreen}{rgb}{0, 153, 76}

\newtcbox{\mybox}[1][blue]{on line, arc=3pt, colback=blue!3!white, fontupper=\ttfamily\small, colframe=blue!20!white, boxsep=0pt, left=1pt, right=1pt, top=2pt, bottom=2pt, boxrule=0.5pt}

\newtcbox{\inlinebox}[1]{%
    on line, %
    arc=2pt, %
    outer arc=2pt, %
    colback=blue!30, %
    colframe=blue!50, %
    boxsep=0pt, %
    left=1mm, right=1mm, top=1mm, bottom=1mm, %
    boxrule=0.5pt, %
    before upper=\strut, %
    fontupper=\strut %
}

\definecolor{darkgreen}{rgb}{0.1,0.5,0.1}

\definecolor{cvprblue}{rgb}{0.21,0.49,0.74}

\usepackage[pagebackref,breaklinks,colorlinks,citecolor=cvprblue]{hyperref}

\def\paperID{10465} %
\def\confName{CVPR}
\def\confYear{2024}

\title{\begin{center}Charting New Territories:\\Exploring the Geographic and Geospatial Capabilities of Multimodal LLMs\end{center}}

\author{%
  Jonathan Roberts \\
  CAML Lab\\
  University of Cambridge\\
  \tt\small{jdr53@cam.ac.uk} \\
  \and
  Timo L{\"u}ddecke \\
  Institute of Computer Science \\
  University of G{\"o}ttingen \\
  \tt\small{timo.lueddecke@uni-goettingen.de}
  \and
  Rehan Sheikh \\
  CAML Lab\\
  University of Cambridge\\
  \tt\small{rs2126@cam.ac.uk} \\
  \and
  Kai Han \\
  Visual AI Lab \\
  The University of Hong Kong \\
  \tt\small{kaihanx@hku.hk} \\
  \and
  Samuel Albanie \\
  CAML Lab\\
  University of Cambridge\\
  \tt\small{samuel.albanie.academic@gmail.com} \\
}

\begin{document}

\twocolumn[{%
\renewcommand\twocolumn[1][]{#1}%
\maketitle

\vspace{-1.6cm}
\begin{center}
  \centering
  \captionsetup{type=figure}

    \begin{subfigure}[b]{0.14\textwidth}
        \includegraphics[height=2.8cm]{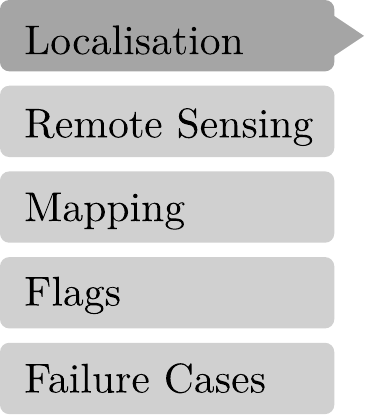}
        \caption{Task overview.}
        \label{fig:geoguess_a}
    \end{subfigure}
    \begin{subfigure}[b]{0.39\textwidth}
        \includegraphics[height=3cm]{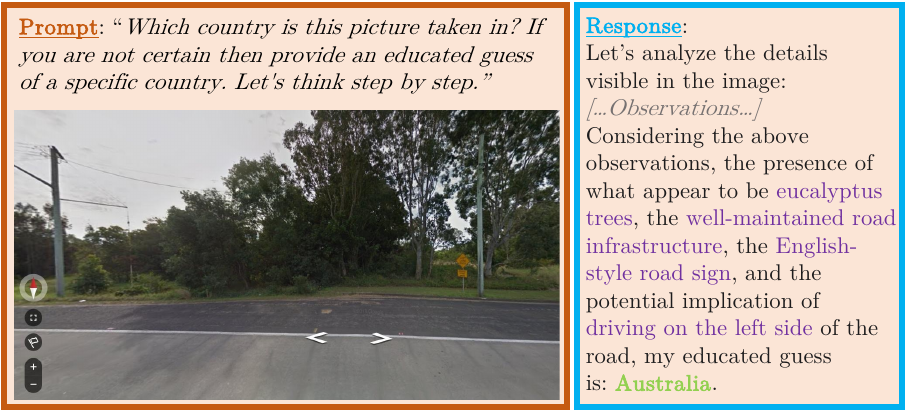}
        \caption{Example prompt and GPT-4V response.}
        \label{fig:geoguess_a}
    \end{subfigure}
    \begin{subfigure}[b]{0.30\textwidth}
        \includegraphics[height=3cm]{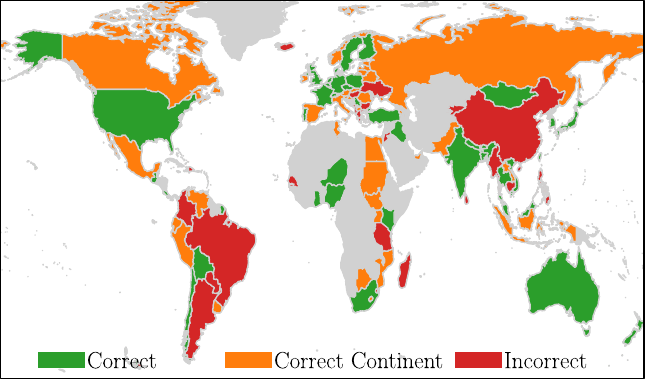}
        \caption{GPT-4V predication accuracy.}
        \label{fig:geoguess_b}
    \end{subfigure}
    \begin{subfigure}[b]{0.12\textwidth}
        \includegraphics[height=3cm]{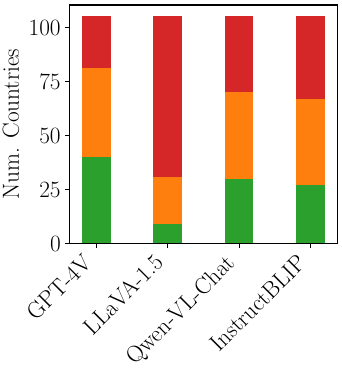}
        \caption{Overall results.}
        \label{fig:geoguess_c}
    \end{subfigure}
    \vspace{-0.2cm}
    \caption{\textbf{Evaluating the geographic and geospatial capabilities of MLLMs.}
    (a) To probe MLLM knowledge we conduct a series of visual experiments. 
    \textit{Localisation}--(b) We pass each image to the MLLM together with the given prompt. As highlighted by its response, we find that GPT-4V is able to extract and reason over fine-grained details in such images. (c) GPT-4V attains promising accuracy on the benchmark across tested countries, where (d) it outperforms other strong MLLM baselines.}
    \label{fig:geoguessr}
\end{center}%
}]

\maketitle

\begin{abstract}
\vspace{-0.75cm}
\vspace{0.15cm}
Multimodal large language models (MLLMs) have shown remarkable capabilities across a broad range of tasks but their knowledge and abilities in the geographic and geospatial domains are yet to be explored, despite potential wide-ranging benefits to navigation, environmental research, urban development, and disaster response. 
We conduct a series of experiments exploring various vision capabilities of MLLMs within these domains, particularly focusing on the frontier model GPT-4V, and benchmark its performance against open-source counterparts. 
Our methodology involves challenging these models with a small-scale geographic benchmark consisting of a suite of visual tasks, testing their abilities across a spectrum of complexity. 
The analysis uncovers not only where such models excel, including instances where they outperform humans, but also where they falter, providing a balanced view of their capabilities in the geographic domain. 
To enable the comparison and evaluation of future models, our benchmark will be publicly released.\footnote{Dataset available at \url{https://github.com/jonathan-roberts1/charting-new-territories}.}
\end{abstract}  

\section{Introduction}
\label{sec:intro}

Driven by key methodological advances (chiefly the Transformer architecture \cite{vaswani2017attention}) and larger-scale datasets and compute infrastructure, vision and language research are currently enjoying a golden age of advancement and progress. The growing capabilities of large language models (LLMs) \cite{bubeck2023sparks, gpt4} have enabled impressive emergent abilities across many different domains. Resultantly, LLMs are becoming ever more prevalent in scientific literature, as well as in wider society. A developing branch of LLM research seeks to widen the spectrum of achievable tasks by incorporating additional modalities, in particular, the visual modality. %
Notable multimodal large language models (MLLMs) include PaLM-E \cite{driess2023palm}, Flamingo \cite{alayrac2022flamingo}, LLaVA-1.5 \cite{liu2023improved}, InstructBLIP \cite{dai2023instructblip}, IDEFICS \cite{laurençon2023obelics}, Qwen \cite{bai2023qwen}, Kosmos-2 \cite{peng2023kosmos}, and recently, GPT-4V \cite{gpt4v}. These models have shown promising potential demonstrating OCR-free mathematical reasoning \cite{driess2023palm}, diagram and document reasoning \cite{yang2023dawn}, and some capacity in medical image interpretation \cite{lee2023cxrllava, wu2023gpt4vision}. While there is a growing body of literature examining the geographic and geospatial capabilities of pure language models \cite{manvi2023geollm, mai2023opportunities} -- including \cite{roberts2023gpt4geo}, which provides an investigation into the factual and application-centric geographic capabilities of LLMs -- there is yet to be such study of multimodal LLMs.

A comprehensive understanding of the geographic capabilities of MLLMs has the potential for a plethora of downstream societal and research applications. The ability to interpret natural images and maps could aid navigation, routing and localisation, and help answer complex questions. Similarly, understanding remote sensing data -- \textit{ex-situ} data derived from air or spaceborne platforms -- could empower many research areas, particularly environmental research and disaster response. Conducting an evaluation of the geographic capabilities of MLLMs is especially timely as the current generation of state-of-the-art models -- and principally, the recent release of frontier model GPT-4V \cite{gpt4v} -- have surpassed a critical threshold where they have sufficient capabilities to understand and at least attempt to perform most complex visual tasks.

However, due to the breadth of the abilities of large models, and the nature of emergent abilities \cite{wei2022emergent}, it is difficult to know \textit{a priori} the tasks a specific model can perform. A corollary to this is the added complexity of model stochasticity, which makes a comprehensive characterisation of ability more problematic. These difficulties are exacerbated by the closed-source nature of some models and the challenges associated with understanding the distribution of data within a pretraining set on the scale of billions or trillions of tokens.
These complexities necessitate exploration by experimentation in a manner that more closely resembles inductive rather than deductive reasoning: deriving general hypotheses and heuristics regarding MLLM performance on different tasks based on mostly qualitative observations.

To this end, we devise and carry out a set of visual geographic experiments that probe the key geographic and geospatial knowledge and reasoning abilities of a suite of MLLMs in both qualitative and quantitative settings, with a focus on determining the extent of the abilities of the frontier model, GPT-4V. To aid reproducibility and facilitate future model comparisons and benchmarking, we release a small-scale benchmark of our experiments.
This dataset partially contains newly generated samples which were not crawled from the Internet.
This strategy helps to mitigate test set contamination, in which models encounter evaluation instances during training.
Finally, we distill our findings into the following \textbf{key takeaways} to inform the research community:
\begin{itemize}
    \item Of all the evaluated models, GPT-4V %
    can perform the broadest range of tasks.
    However, it does not always perform best, \eg, satellite image detection and classification tasks.
    In general, it recognises fine-detail well but tends to fail when precise localisation is required.
    \item More broadly, the best model choice depends on the task at hand. Qwen-VL and LLaVA-1.5 in particular often demonstrate good localisation performance.
    \item Enforcing a specific output format is challenging, models often resort to explanations why they are not capable of performing the task. Among the evaluated models GPT-4V was least susceptible to this behaviour.
    \item The current generation of leading MLLMs suffer a performance penalty when processing multi-object images, relative to their performance on single object images.
\end{itemize}

\section{Related Work}
\label{sec:related-work}
\noindent\textbf{Geographic Capabilities of LLMs.}
A comprehensive study of the performance of various LLMs on different geospatial intelligence tasks in given by \cite{mai2023opportunities}. Other works investigate more specific tasks, such as the extraction of geospatial knowledge \cite{manvi2023geollm} or as a mapping assistant \cite{juhasz2023chatgpt}. GPT4GEO \cite{roberts2023gpt4geo} includes a mixture of qualitative and quantitative evaluations of the geographic capabilities of GPT-4 across various knowledge-based and reasoning experiments. We take inspiration from \cite{roberts2023gpt4geo}, building upon it by (1) expanding our experiments to include visual inputs; (2) including comparisons to open-source models; and (3) releasing a small-scale experimental benchmark.

\noindent\textbf{Geographic Capabilities of MLLMs.}
There is yet to be an extensive evaluation of MLLM capabilities in the geographic domain. However, various geographic or geospatial examples can be found in more general evaluations. In \cite{mai2023opportunities}, the performance of OpenFlamingo-9B on a remote sensing classification dataset is evaluated. %
A handful of satellite image descriptions from GPT-4V and Bard are evaluated in \cite{Noever2023multimodal}. The broad survey conducted in \cite{yang2023dawn} includes a small number of samples of GPT-4V identifying landmarks, foods and interpreting maps. Our work includes all of these tasks to a greater depth as well as introducing many others, including particular attention to the interpretation of remote sensing data, a crucial data source for vision research.

\noindent\textbf{Other Capabilities of MLLMs.}
Traditional evaluations of MLLMs are carried out as defined benchmarks, such as MME \cite{fu2023mme}, SEED-Bench \cite{li2023seed} and MMBench \cite{liu2023mmbench}, totaling 1000s of questions and corresponding answers. However, the recent release of GPT-4V has sparked a flurry of more qualitative and open-ended evaluations that probe capability beyond a simple question-answer setting. Examples of such works for specific tasks include using GPT-4V for embodied decision-making \cite{chen2023towards}, for self-driving \cite{wen2023road} and various medical applications \cite{lee2023cxrllava, wu2023gpt4vision}. More general studies, such as \cite{yang2023dawn} includes a broad collection of intriguing general vision examples and \cite{wu2023early}, which also includes interpretation of infrared and depth imagery, as well as mel spectograms. We build upon these preliminary analyses of \cite{yang2023dawn} and conduct a detailed exploration focused on the capabilities of GPT-4V in the geographic domain while adding quantitative experiments and comparisons to other MLLMs.

\section{Methodology}
\label{sec:methods}

\subsection{Experimental Design}
In designing our experiments, we draw inspiration from \cite{roberts2023gpt4geo} and create a suite of both qualitative and quantitative experiments. We categorise our experiments according to the type of visual input and include geographic tasks covering natural (photographs), abstract (\eg, maps/flags), and remotely sensed (satellite) images. 
As large language models are trained on larger web-crawled datasets, an increasing problem for evaluation is data contamination \cite{magar-schwartz-2022-data}, i.e. test data has been seen during training. Although we leverage established datasets in some experiments, the majority of our evaluation is  conducted on visual inputs that are sufficiently curated to make their existence in the model training distribution highly unlikely.

We adopt this approach of providing a broad capability overview instead of a systematic study, for two main reasons:
(1) at the time of writing, the GPT-4V API is rate-limited to 100 requests a day, too few for a large-scale study, 
(2) an automated evaluation would require structured output, however, current models cannot reliably be controlled for this when pushed close to the limits of their abilities.
Consequently, it was important to adopt a methodology that could deliver insights despite these limitations.

\subsection{Models}
We focus our experimentation on the current most capable model, GPT-4V \cite{gpt4v}, and query it through both the ChatGPT interface\footnote{\url{https://chat.openai.com/}} and API\footnote{\url{https://openai.com/blog/openai-api}}. As a comparison, we also evaluate the open-source LLaVA-v1.5-13b \cite{liu2023improved}, IDEFICS-80b-Instruct \cite{laurençon2023obelics}, Qwen-VL-Chat \cite{bai2023qwen}, InstructBLIP-Vicuna-13b \cite{dai2023instructblip} and Kosmos-2 \cite{peng2023kosmos} models, accessed predominantly through HuggingFace Transformers \cite{wolf2020huggingfaces}. See \hyperref[app]{Appendix} for specific model hyperparameters. Hereafter, references to these models refer to the specific models mentioned here.

\subsection{Prompting}
Given the wide range and unchartered nature of our experiments, we treat prompting as a hyperparameter that we tune for each model and experiment. We leverage prompt engineering techniques (\eg, 0-shot CoT \cite{kojima2023large}) where relevant and discuss their impact, however, a systematic comparison of strategies is not appropriate for this setting. %

\section{Experiments}
\label{sec:experiments}

\subsection{Localisation}
Accurately inferring the location an image is taken has applications ranging from forensic analysis to navigation in GPS-denied environments.
To probe this capability, we take inspiration from the popular geography game, \textit{GeoGuessr\footnote{\url{https://www.geoguessr.com/}}}, in which players are placed somewhere in the world in a street view panorama and have to guess their location. We construct a dataset of 100 images by randomly sampling a single image for each sovereign state in the GeoLocation dataset \cite{kaggle_geoguessr}.
Using the prompt shown in Fig. \ref{fig:geoguessr}, we query each model to predict the country each image was taken in. We find that GPT-4V in particular is able to extract small details from the images (such as species of foliage, road signs, advertised products, and architecture) and reason over them when making a prediction -- see Fig. \ref{fig:geoguess_a}. In this experiment, we do not detect strong geographic biases from GPT-4V, although countries in S. America were frequently mistaken for other continents (Fig. \ref{fig:geoguess_b}). Though inferior to GPT-4V, Qwen-VL and InstructBLIP performed well (Fig. \ref{fig:geoguess_c}); on the other hand, a proclivity to overpredict the USA prevented LLaVA-1.5 from scoring well.

{
\setlength{\parskip}{-2pt}

\subsection{Remote Sensing}
The ability to interpret remotely sensed (RS) data enables numerous positive applications including environmental research and land-use planning. In the following section, we outline a series of tasks involving RS data that test capabilities ranging from holistic image classification and understanding to fine-grained detail extraction and localisation.

\paragraph{Classification.} As current MLLMs are limited to pure language output, we initially explore their ability classify RS imagery with image-level labels. To broadly characterise this ability in the RS domain we evaluate performance on a subset of the SATIN metadataset \cite{roberts2023satin}, a challenging benchmark that includes %
different resolutions, fields of view sizes, class categories, and imagery types. We take a 12-image class-balanced sample from two randomly selected datasets from each of the 6 tasks in the benchmark (totaling 144 images).
The zero-shot classification results on this subset are shown in Tab. \ref{table:satin_results}. There is some variation in ranking across the different tasks, though overall LLaVA-1.5 performs best. A mean accuracy of 0.56 for LLaVA-1.5 might seem low; however, this represents the challenge of the benchmark -- vision-language baselines \eg CLIP \cite{radford2021learning} score comparably on the overall SATIN benchmark \cite{roberts2023satin}. A necessary caveat to this comparison is that we only evaluate a narrow subset of the benchmark. %
To partly address this, we evaluate LLaVA-1.5 on 4 additional subsets of the data (resampled with replacement) and report a low variance, giving us confidence that the subset used for evaluating the other models is representative.

}

\begin{table}
\centering
\footnotesize
\setlength{\tabcolsep}{2pt}
\begin{tabular}{llllllllll}
\toprule
          & \multicolumn{6}{c}{\textbf{Zero-Shot Classification Accuracy \textit{per task}}} &                                                           \\

\textbf{\textit{Model}} & \textit{1} & \textit{2} & \textit{3} & \textit{4} & \textit{5} & \textit{6} & \textit{\textbf{Mean}} \\ \hline
\textbf{GPT-4V}    & \textbf{0.50}   & 0.67            & \textbf{0.57}   & \textbf{0.53}   & 0.46            & 0.38            & 0.52          \\
\textbf{LLaVA-1.5} & 0.42$\pm$\light{0.2}            & \textbf{0.71}$\pm$\light{0.1}    & 0.50            & 0.48            & \textbf{0.71}$\pm$\light{0.1}    & \textbf{0.54}    & \textbf{0.56}  \\
\textbf{Qwen-VL}   & 0.25            & 0.63            & 0.40            & 0.37            & 0.50            & 0.38            & 0.42    \\
\bottomrule
\end{tabular}
\vspace{-0.3cm}
\caption{\textbf{0-shot satellite imagery classification} on a subset of SATIN \cite{roberts2023satin}. For LLaVA-1.5 we report $\mu\pm\light{\sigma}$ across 5 subsets.}
\label{table:satin_results}
\end{table}

\begin{figure*}[t]
    \centering
    \includegraphics[width=0.9\textwidth]{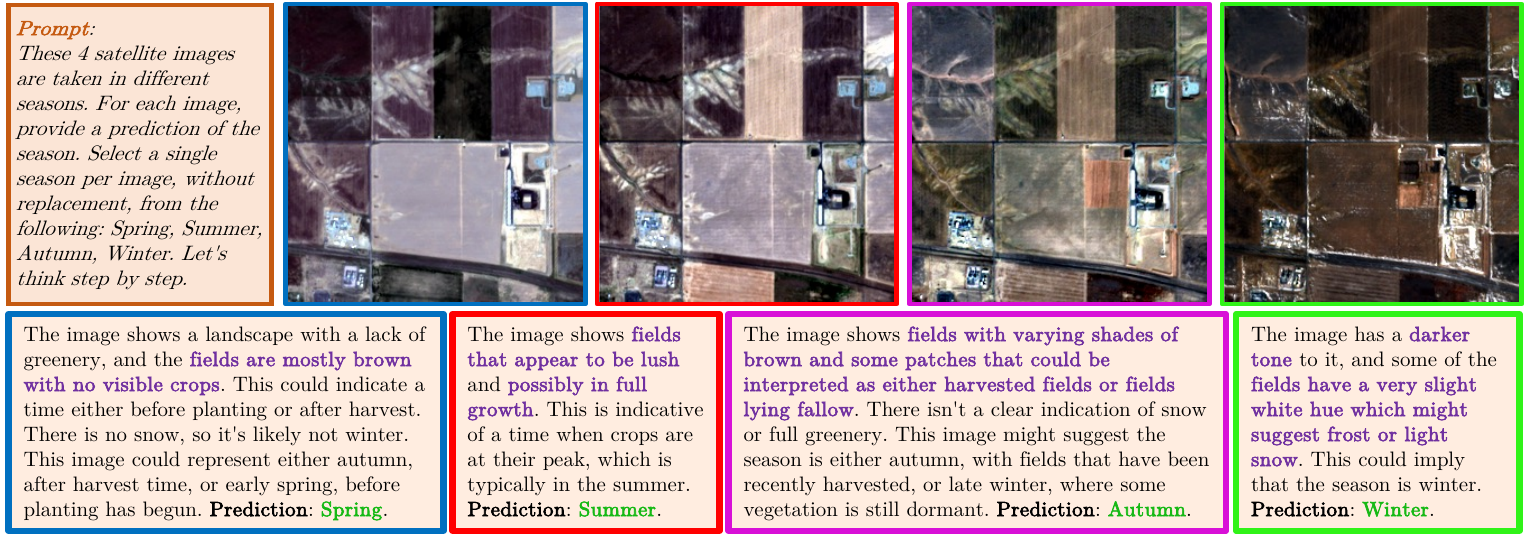}
    \vspace{-0.1cm}
    \caption{\textbf{Satellite imagery change detection}. We test GPT-4V's ability to detect seasonal changes in a four-image time-series from \cite{wang2023ssl4eos12}. In this example, the model is able to pick up minor details such as crop colouration and the presence of snow to correctly estimate seasons.}
    \label{fig:time-series}
\end{figure*}

\paragraph{Change Detection.} A key usage of RS data is monitoring and more specifically, \textit{detecting change}. The ability to identify fine-grained differences in an image time series and conduct a nuanced analysis of the cause has many downstream uses. It is not clear the extent to which current MLLMs can perform this task, though preliminary investigations, \eg, \cite{wu2023early} suggest they struggle. We conduct a small-scale experiment probing the degree to which GPT-4V can determine the season of images in time series from \cite{wang2023ssl4eos12}. We select 6 four-season quadlets (same location) that are clear with minimal atmospheric effects and evaluate GPT-4V's interpretation. This task is particularly challenging as the imagery contains only minor differences that are difficult to attribute to a particular season, hence GPT-4V scores $\sim$38\% accuracy. However, this relatively low accuracy is not due to an inability to detect changes, but rather to their interpretation. Fig. \ref{fig:time-series} illustrates GPT-4V's correct analysis of one of the quadlets, demonstrating good detection of fine-grained details and reasoning. We conclude that GPT-4V displays impressive, though fledgling capabilities to interpret fine-grained details in time-series RS imagery.

\paragraph{Segmentation.} Having demonstrated the ability to describe fine-grained details in images in the previous experiment, we increase the task complexity to segmentation: identifying \textit{and} localising objects and classes. Despite not being able to provide visual outputs, it is possible -- with creative prompting -- to create segmentation maps from pure language. We introduce two such prompting strategies: (1) \textbf{Grid Segmentation}: \hl{Segment the image into the following [classes]. Display the results as a [YxY] table with each cell labelled with one of the class labels}, and (2) \textbf{SVG Segmentation}: \hl{Segment the image into the following [classes]. Provide the code for an SVG that displays the segmentation map}.

\begin{figure*}[ht]
    \centering
    \begin{subfigure}[b]{0.17\textwidth}
        \includegraphics[width=\textwidth]{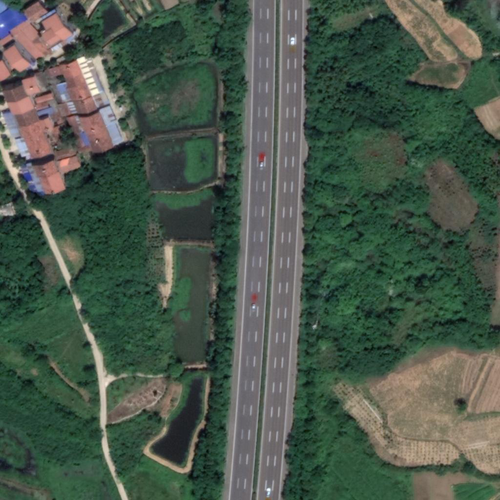}
        \caption{Image}
        \label{fig:first}
    \end{subfigure}
    \begin{subfigure}[b]{0.17\textwidth}
        \includegraphics[width=\textwidth]{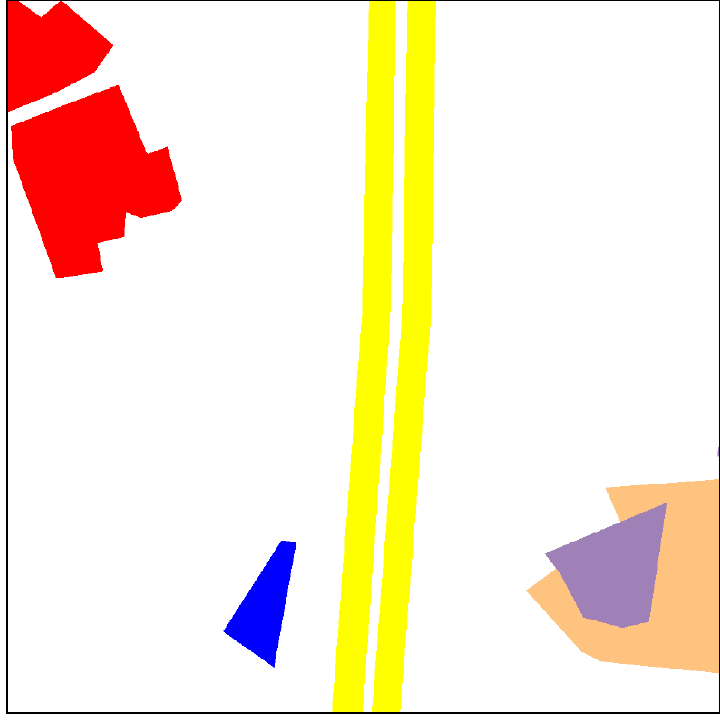}
        \caption{Ground Truth}
        \label{fig:second}
    \end{subfigure}
    \begin{subfigure}[b]{0.17\textwidth}
        \includegraphics[width=\textwidth]{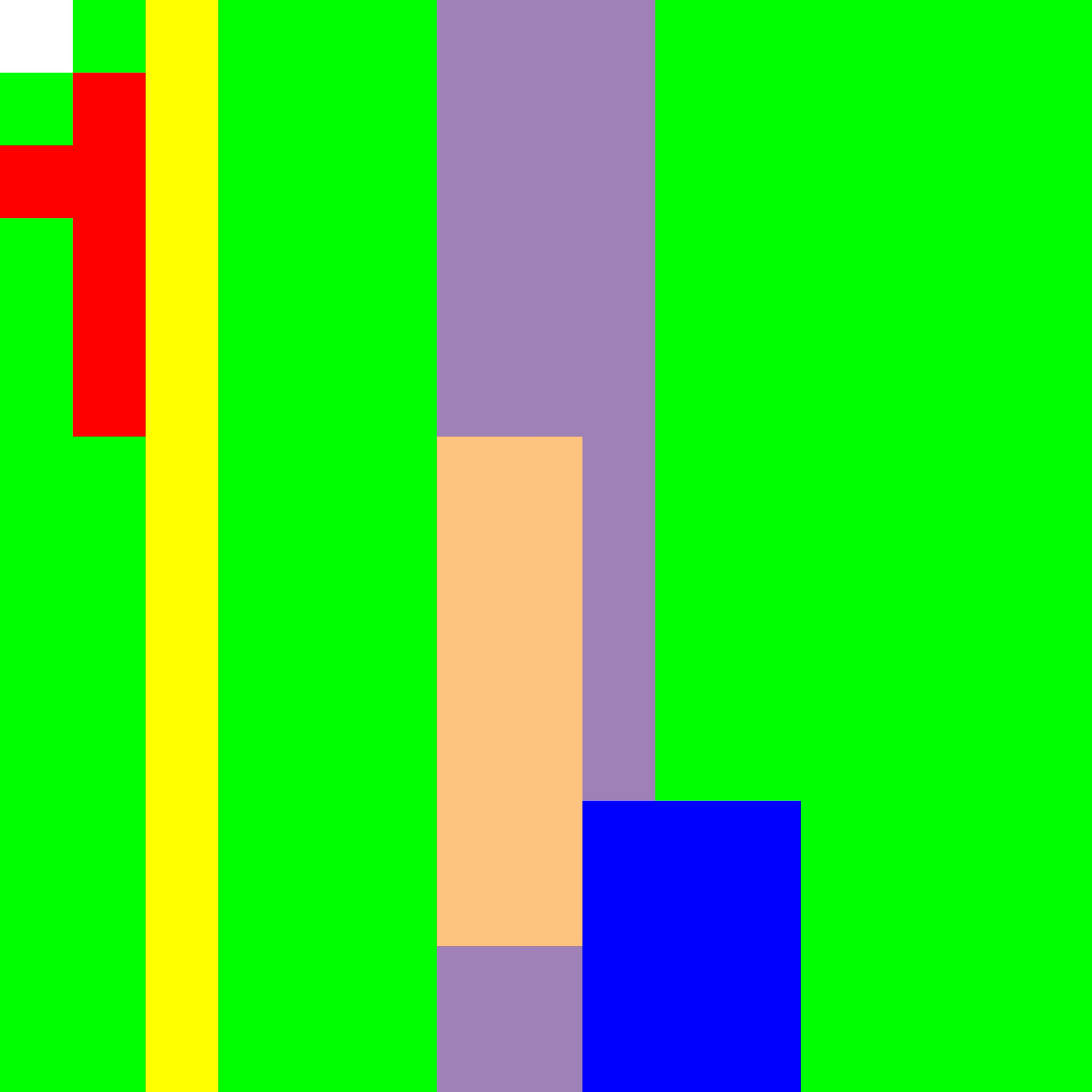}
        \caption{Grid Segmentation}
        \label{fig:third}
    \end{subfigure}
    \begin{subfigure}[b]{0.17\textwidth}
        \includegraphics[width=\textwidth]{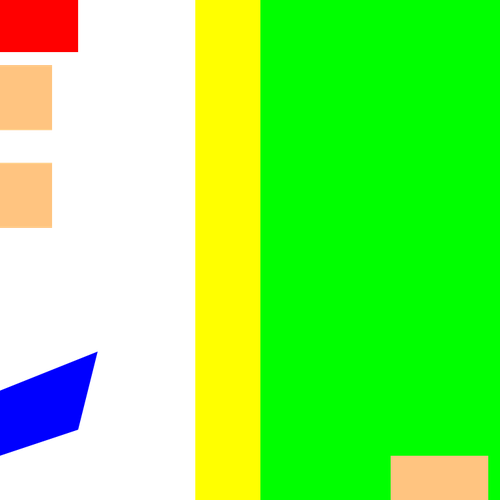}
        \caption{SVG Segmentation}
        \label{fig:fourth}
    \end{subfigure}
    \begin{subfigure}[b]{0.17\textwidth}
        \includegraphics[width=\textwidth]{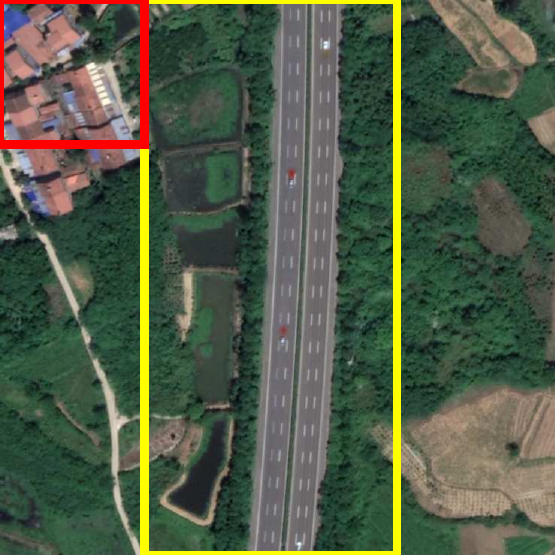}
        \caption{Bounding Boxes}
        \label{fig:fourth}
    \end{subfigure}
    \begin{subfigure}[b]{0.10\textwidth}
        \includegraphics[width=\textwidth]{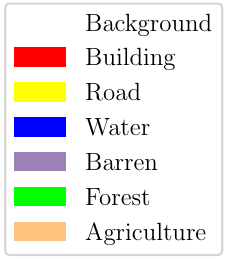}
        \caption{Classes}
        \label{fig:fifth}
    \end{subfigure}

    \vspace{-0.1cm}
    \caption{\textbf{Segmentation using GPT-4V}. We include examples of Grid (c) and SVG (d) segmentation, and localisation (e) of satellite imagery (a) from LoveDA \cite{wang2022loveda}. Bounding boxes are for \colorbox{gray!50}{\textcolor{red}{urban areas}} and \colorbox{gray!50}{\textcolor{yellow}{road}}. Segmentation labels are given in (f).}
    \label{fig:segmentation}
\end{figure*}

We include qualitative examples of each prompting strategy in Fig. \ref{fig:segmentation}. Clearly, the segmentation maps are not perfect and underperform vision-language or supervised vision baselines; however, a number of positive observations can be made. Generally, the relevant classes are detected and included in the map. Additionally, key features such as building groups, roads, and water bodies are detected and localised in approximate positions. Furthermore, annotation inaccuracies are demonstrated as the ground truth in the top row is mostly labelled as background but perhaps more appriopriately given the forest label suggested by GPT-4V. Considering a broader range of examples, we observe grid segmentation to outperform SVG segmentation. We find that there is not sufficient capacity in the other MLLMs to perform segmentation.

\setlength\fboxsep{2pt}

\begin{figure*}
    \centering
\begin{minipage}{.5\textwidth}
\centering
    \begin{subfigure}{0.32\textwidth}
        \centering
        \includegraphics[height=2.2cm]{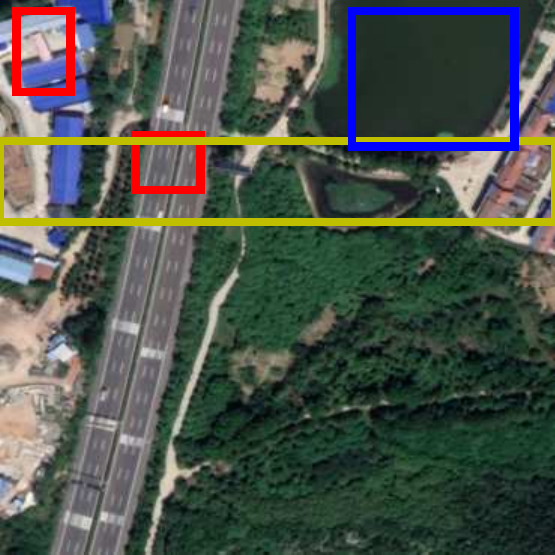}
        \caption{GPT-4V}
    \end{subfigure}
    \begin{subfigure}{0.32\textwidth}
        \centering
        \includegraphics[height=2.2cm]{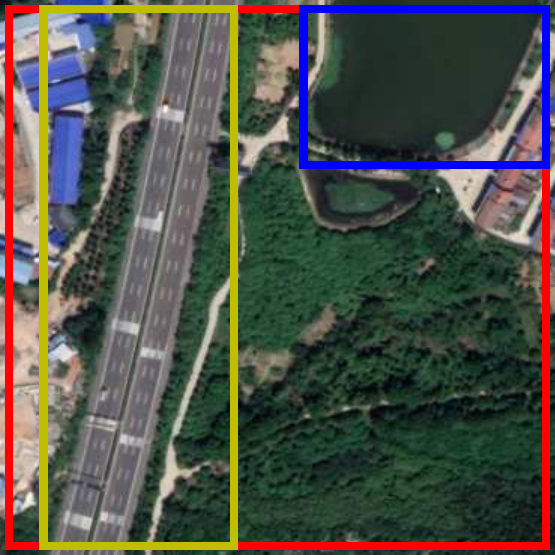}
        \caption{Kosmos-2}
    \end{subfigure}
    \begin{subfigure}{0.32\textwidth}
        \centering
        \includegraphics[height=2.2cm]{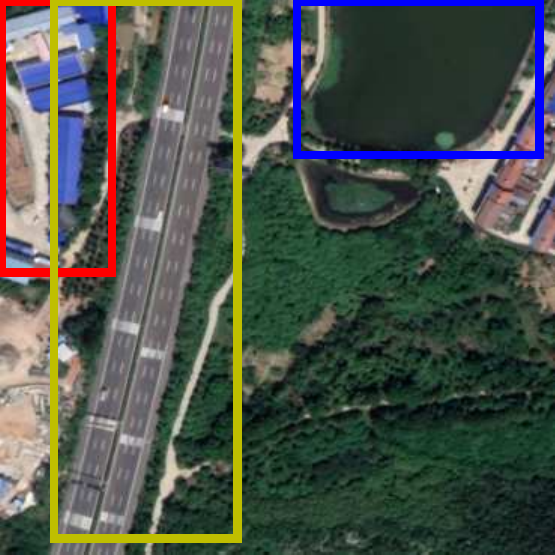}
        \caption{Qwen-VL}
    \end{subfigure}
    \newline
    \begin{subfigure}{0.32\textwidth}
        \centering
        \includegraphics[height=2.2cm]{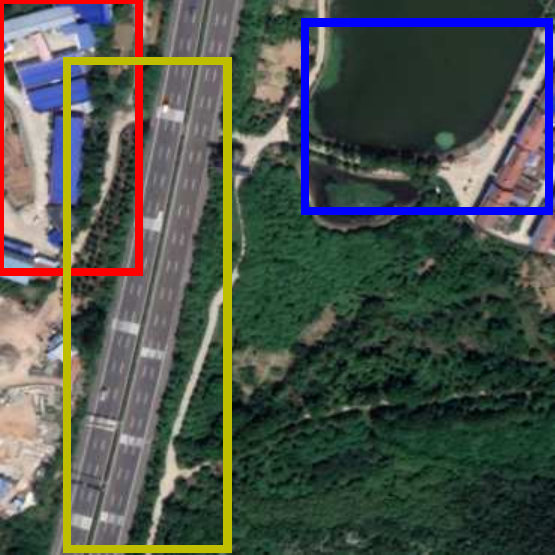}
        \caption{LLaVA-1.5}
    \end{subfigure}
    \begin{subfigure}{0.32\textwidth}
        \centering
        \includegraphics[height=2.2cm]{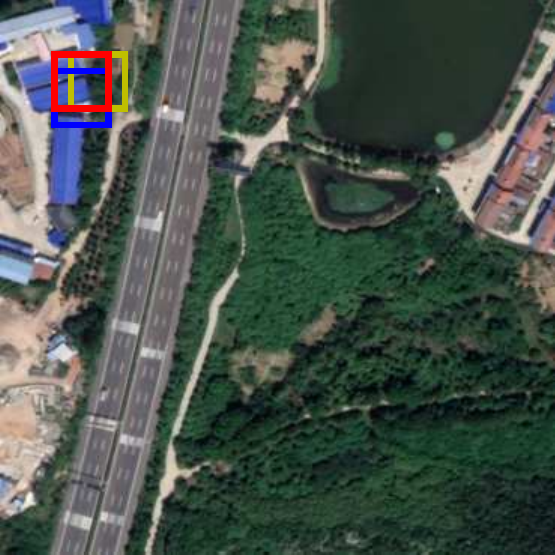}
        \caption{IDEFICS$^*$}
    \end{subfigure}
       \begin{subfigure}{0.32\textwidth}
        \centering
        \footnotesize{$^*$ all IDEFICS bounding boxes had the same coordinates. A minor offset has been applied for visualisation.}\\
        \caption*{}
    \end{subfigure}
    \vspace{-0.2cm}
    \caption{\textbf{Bounding boxes} for \colorbox{gray!50}{\textcolor{red}{urban areas}}, \colorbox{gray!50}{\textcolor{yellow}{road}} and \colorbox{gray!50}{\textcolor{blue}{water bodies}}.}
    \label{fig:bounding_boxes}
\end{minipage}%
\hfill
\begin{minipage}{.5\textwidth}
\centering
\includegraphics[height=5.3cm]{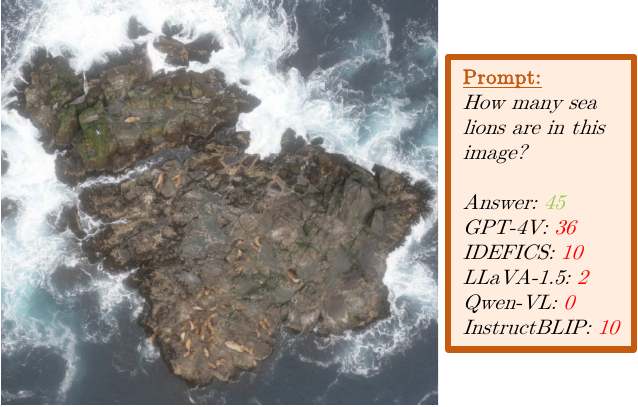}
\vspace{-0.2cm}
        \caption{\textbf{Counting} small objects proves challenging.}
        \label{fig:sealions}
\end{minipage}
\end{figure*}

\paragraph{Bounding Boxes.} We also analyse the object detection capabilities of the MLLMs in the form of bounding boxes. We draw a number of qualitative observations from the example bounding boxes in Fig. \ref{fig:bounding_boxes}. As reported in \cite{yang2023dawn}, we find that bounding boxes provided by GPT-4V are mostly inaccurate. Qwen-VL produces tight, accurate boxes that are more accurate than LLaVA-1.5. Kosmos-2 also produces accurate bounding boxes and demonstrates a different -- though arguably also correct -- interpretation of `urban area'. The bounding box from IDEFICS is especially poor.

\paragraph{Counting.} Counting small objects in images probes the resolution of the visual component of each MLLM. In Fig. \ref{fig:sealions}, we include an aerial scene from \cite{kaggle_sealion} comprising 45 sea lions of various sizes lounging on a wave-beaten rock of similar colouration to the sea lions. This particular task is difficult -- GPT-4V's prediction suggests it is close to the limits of what it can perceive, while it is clearly beyond the capabilities of the other models.

\begin{table}
\centering
\footnotesize
\setlength{\tabcolsep}{2pt}
\begin{tabular}
{lccccccccc}
\toprule
 & \multicolumn{7}{c}{\textbf{Mean Distance Error (km)}} \\

 & & & & \textit{North} & & \textit{South} \\
\textbf{Projection}   & \textit{Africa} & \textit{Asia}  & \textit{Europe} & \textit{America} & \textit{Oceania} & \textit{America} & \textit{\textbf{Mean}}  \\  \hline
Mercator     & 3542 & 2486 & 1351 & 2809        & 1657  & 3460        & 2551 \\
Miller       & 9398 & 3391 & 990  & 3334        & 2071  & 7434       & 4436 \\
Mollweide    & 8441 & 2985 & 1289 & 2587        & 2798  & 2998        & 3516 \\
PlateCarree  & 9339 & 2730 & 1874 & 3032        & 2058  & 3215        & 3708 \\
Robinson     & 2796 & 2501 & 1040 & 3661        & 1984  & 3230        & \textbf{2535} \\
\midrule
\textbf{Mean}         & 6703 & 2819 & \textbf{1309} & 3084        & 2114  & 4067        & 3349 \\
\bottomrule
\end{tabular}
\vspace{-0.3cm}
\caption{\textbf{Localisation: map $\rightarrow$ real-world.} The mean distance error is calculated over 10 points on each continent map crop.}
\label{tab:projections}
\end{table}

\subsection{Mapping} %
The ability to understand and interpret maps is a core geospatial skill required for route planning, urban development, and disaster response. The fundamental requirement for successful map reading is relating areas and positions on maps to the real-world locations they represent. To evaluate this capability, we carry out a series of map interpretation experiments that focus on identifying geographic entities from maps, localising points on maps, and finally, map annotation. %

\begin{figure*}[t]
    \vspace{-0.3cm}
    \includegraphics[width=0.27\textwidth]{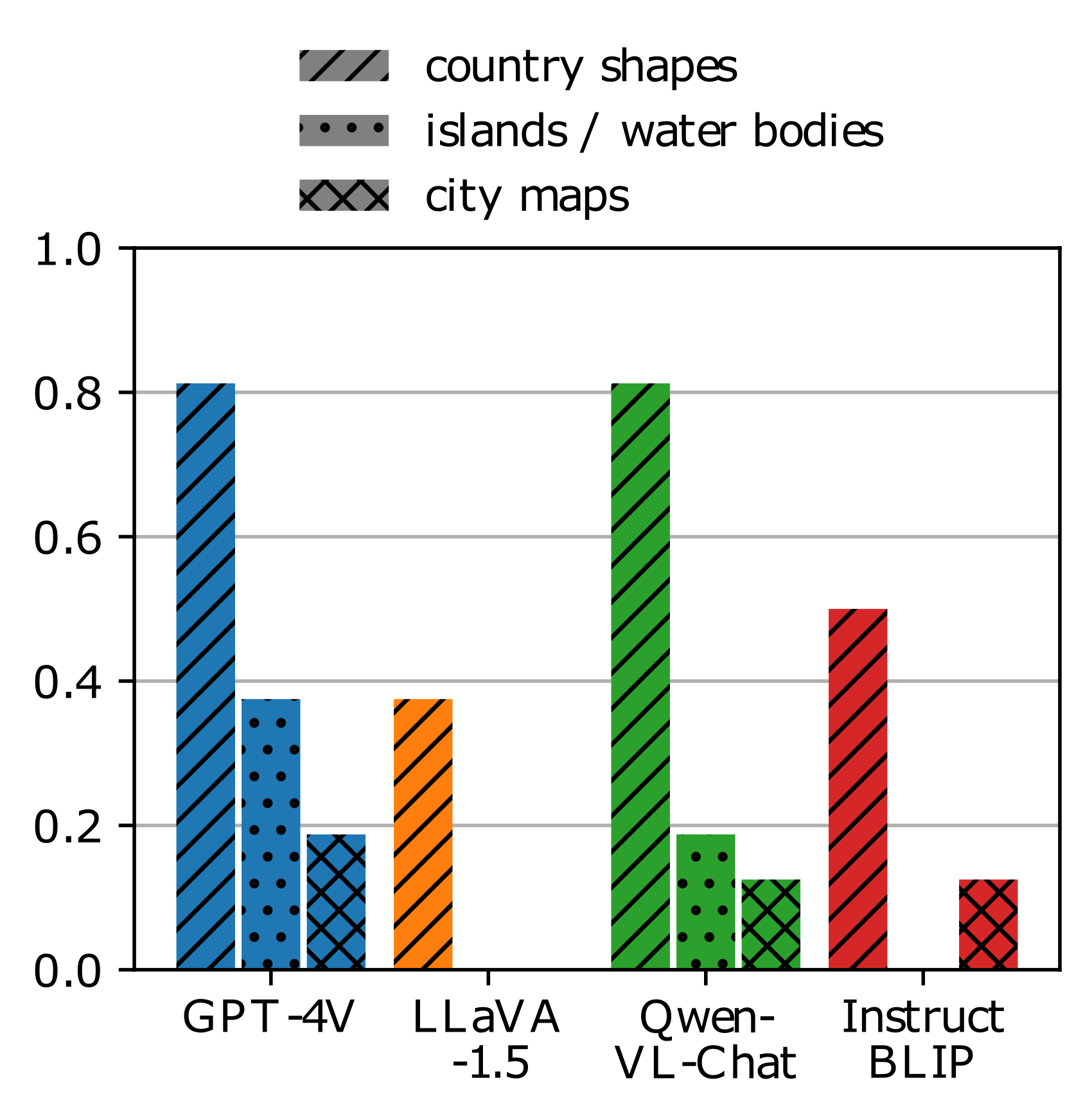}
    \includegraphics[width=0.71\textwidth]{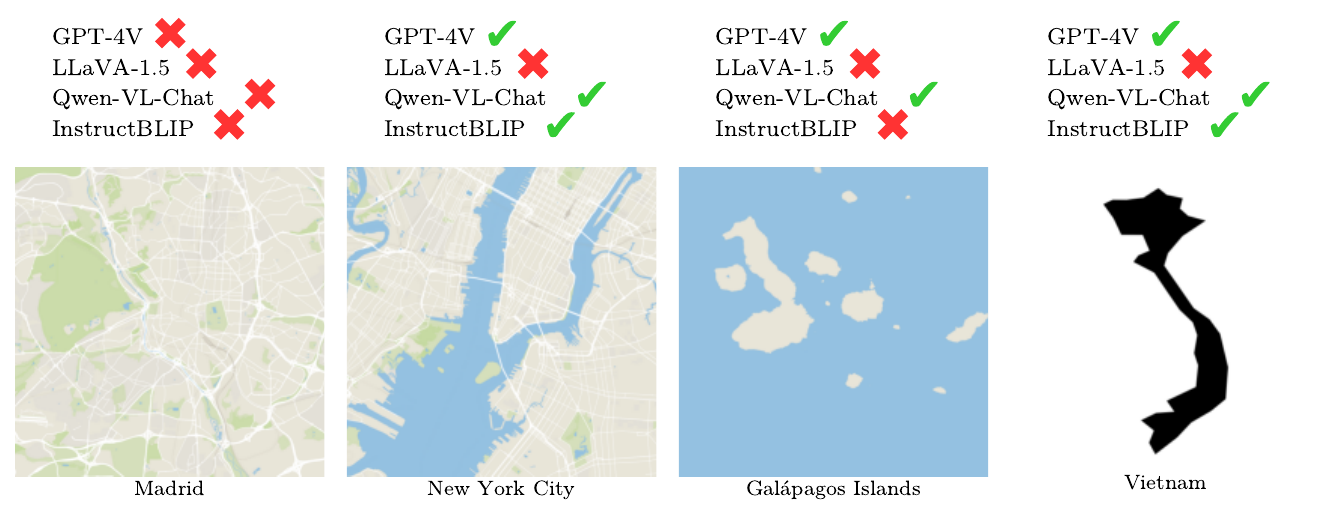}
    \vspace{-0.3cm}
    \caption{\textbf{Identification:} [L]: quantitative results for all three identification tasks. [R]: selected examples for each task indicating success and failure of the individual models (map data from OpenStreetMap). (Note, high resolution images were provided to the models.)}
    \label{fig:abstract_geo}
\end{figure*}

{
\setlength{\parskip}{-3pt}

\paragraph{Region Identification.} We collect a compact dataset of three identification tasks: (A) state from outline shape, (B) island or water body naming, and (C) city from map.
Each task comprises 16 samples generated by the authors, minimising the risk of test-set contamination.
These experiments assess pattern and shape processing capabilities of the vision model on the one hand but also visual world knowledge on the other hand.
We show the results in Fig.~\ref{fig:abstract_geo}.

\paragraph{A. State name from outline}

In the first experiment, the models are tasked with inferring a state's name from the shape of its outline.
To obtain data, we use GeoPandas \cite{jordahl2019geopandas} to render state outlines.
The results show a good performance of GPT-4V and Qwen-VL while InstructBLIP and LLaVA-1.5 perform worse.

\paragraph{B. City name from maps}
Given a city map image, we ask the models to identify the corresponding city.
To generate the data, we use the OpenStreetMap-based Maputnik tool \cite{maputnik} and extract maps at zoom-level 12 at the city locations provided by the GeoNames database \cite{geonames}. Cities are sampled balanced over the continents. We extract images of size $1000 \times 1000$ pixels.
Other models receive single images as prompts.
The responses of GPT-4V suggest it captures the patterns in the map data but lacks the visual knowledge to associate them with specific locations on Earth. Other models perform worse. LLaVA-1.5 (and GPT-4V to a smaller degree) often predict well-known cities like New York or San Francisco indicating a bias towards presumably frequently mentioned cities in the training data. These cities also count among the few samples the models classified correctly, the models were consistent in this.
GPT-4V describes scenes well but appears to fail to retrieve the right name. This could be because information encoded in city maps is rarely expressed in text and only selected city maps were seen during training.
InstructBLIP often outputs nonsensical answers for the provided prompt, highlighting a deficit in instruction following.

\paragraph{C. Island and water body naming from maps}
In the last experiment, islands and water bodies need to be identified.
We use the same method for obtaining data as in the previous city map experiment (B) but select locations and zoom level manually. Furthermore, we simplify the maps to only incorporate land cover and sea, since we found this to work better in early experiments.
The task is easier than understanding city maps but still challenging as even the best model, GPT-4V, correctly classified only 6 out of 16 samples.

}

\begin{figure*}[ht]
  \centering
  \begin{minipage}{.53\textwidth}
  \centering
    \includegraphics[height=4.5cm]{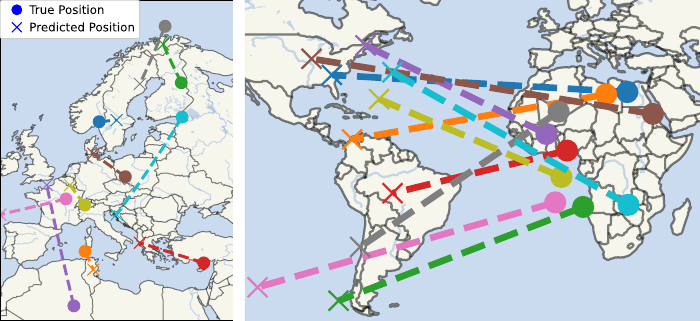}
    \caption{\textbf{Localisation: map $\rightarrow$ real-world}: predicted latitude and longitude positions. [L] Europe (Miller), [R] Africa (PlateCarree).}
    \label{fig:projections_output} %
  \end{minipage}
  \hfill
  \begin{minipage}{.45\textwidth}
  
      \centering
  \includegraphics[height=5cm]{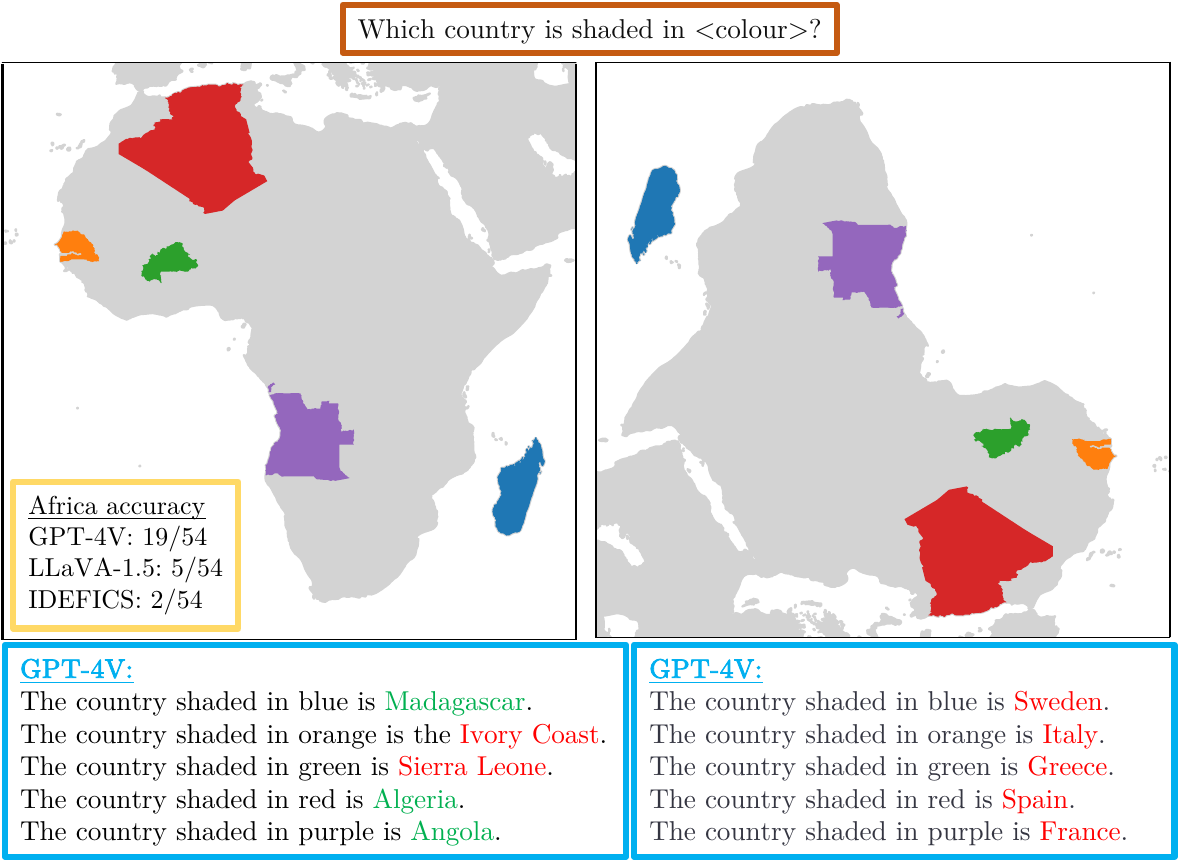}
  \caption{\textbf{Identifying multiple states} proves difficult on a regular map and is a clear failure case when the map is flipped.}
  \label{fig:shaded_africa}
  \label{fig:projections}
  \end{minipage}
\end{figure*}

\paragraph{Localisation: map $\rightarrow$ real-world.}
Next, we look to test the accuracy of the models in predicting the real-world positions (latitude and longitude) of query points on maps. To do so, we create equal latitude and longitude extent map crops (using Cartopy \cite{cartopy}) of each continent and place ten coloured points on each %
We repeat this process to create equivalent maps for 5 common map projections, and prompt GPT-4V with: \hl{\textit{`Estimate the Latitude/Longitude positions of the coloured points on the map'}}, to obtain a position estimate for each point. Using the haversine formula, we calculate the error between the true (circles) and predicted (crosses) positions of each point; the mean position errors for each map are shown in Tab. \ref{tab:projections}. A number of insights can be gleaned from these results, firstly that GPT-4V struggles with the task, with even the most accurate configuration (Europe, Miller) scoring an average error of nearly 1000 km (Fig. \ref{fig:projections_output}). Frequently, GPT-4V confuses the continent the points are placed on, such as in the lowest accuracy setting, where points in/around Africa are predicted in North and South America (Fig \ref{fig:projections_output}). A clear geographic bias is evident with the lowest errors in Europe and significantly higher errors in Africa. %
This is in agreement with other results, such as the difficulties identifying African countries (see \textit{Failure cases}), and plotting Africa's outline in \cite{roberts2023gpt4geo}, potentially suggesting that geographic information about Africa is less prevalent in the training data.
We also observe variation in error amongst the different projections, with the Mercator and Robinson proving the easiest for GPT-4V to interpret. This variation could be caused by factors such as the interpretability of specific projections (especially at extreme latitudes) or the prevalence of particular projections in the training distribution.

\begin{figure*}[ht]
\centering
    \includegraphics[width=0.78\textwidth]{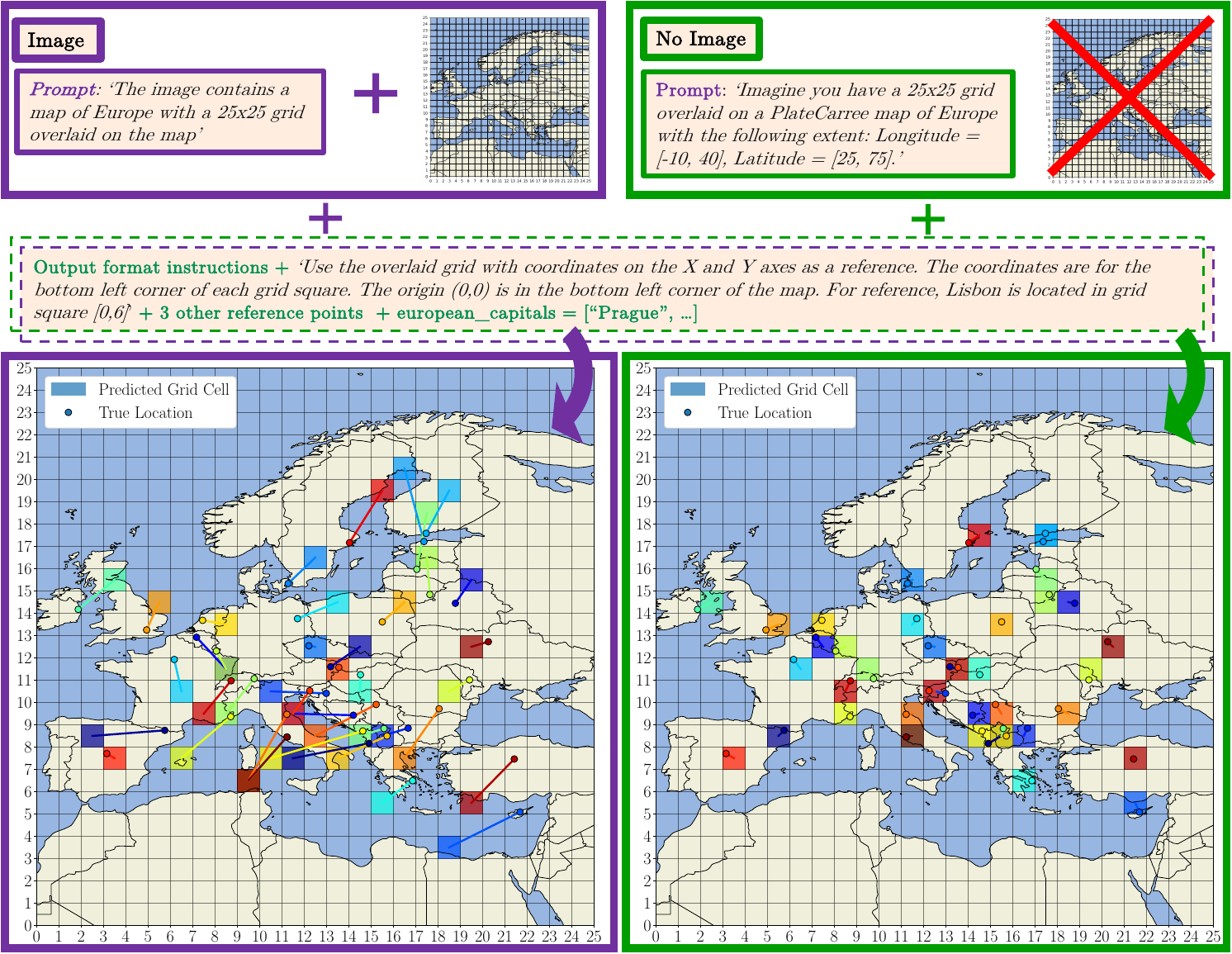}
    \caption{\textbf{Localisation: real-world $\rightarrow$ map}. Using GPT-4V to annotate European capital cities on a map overlaid with a grid in two settings: \textcolor{violet}{\textbf{Image}} [left] where the map grid is passed to the model and \textcolor{darkgreen}{\textbf{No Image}} [right] where just a brief description of the map is passed.}
    \label{fig:grid_coordinates}
\end{figure*}

\paragraph{Localisation: real-world $\rightarrow$ map.}
The final component of the cartographic experimentation tests the ability to localise in the inverse direction: annotating points on the map given a real-world location. Concretely, given the names of European capital cities, we prompt GPT-4V to provide their positions on a bespoke coordinate system overlaid on a map of Europe (created using Cartopy) -- details of the inputs and prompts are illustrated in Fig. \ref{fig:grid_coordinates}. We analyse GPT-4Vs performance on two different task settings: (1) \textbf{Image} in which we pass the map and grid to GPT-4V and (2) \textbf{No Image} in which we just pass a brief description of the map. When shown the map, GPT-4V struggles to accurately locate the capital cities on the grid, despite the grid cells being relatively coarse, locating only 1 capital (Madrid) within the correct grid cell. Conversely, when simply prompted with a description of the map -- including the \textit{extent} -- GPT-4V is able to locate \textit{every} capital in the correct grid cell. It does so by performing the following: (i) estimate the approximate latitude/longitude of each capital using its knowledge, (ii) determine a transformation from latitude/longitude to the grid reference, and (iii) perform the conversion. This dichotomy in performance highlights GPT-4V's strong capabilities to solve complex problems in a pure language setting while struggling to perform precise and accurate image analysis. Performance varies depending on the grid resolution.

\subsection{Flags}
\label{subsec:flags}

\begin{figure*}[ht]
\centering
\begin{subfigure}[b]{0.38\textwidth}
\includegraphics[width=\textwidth]{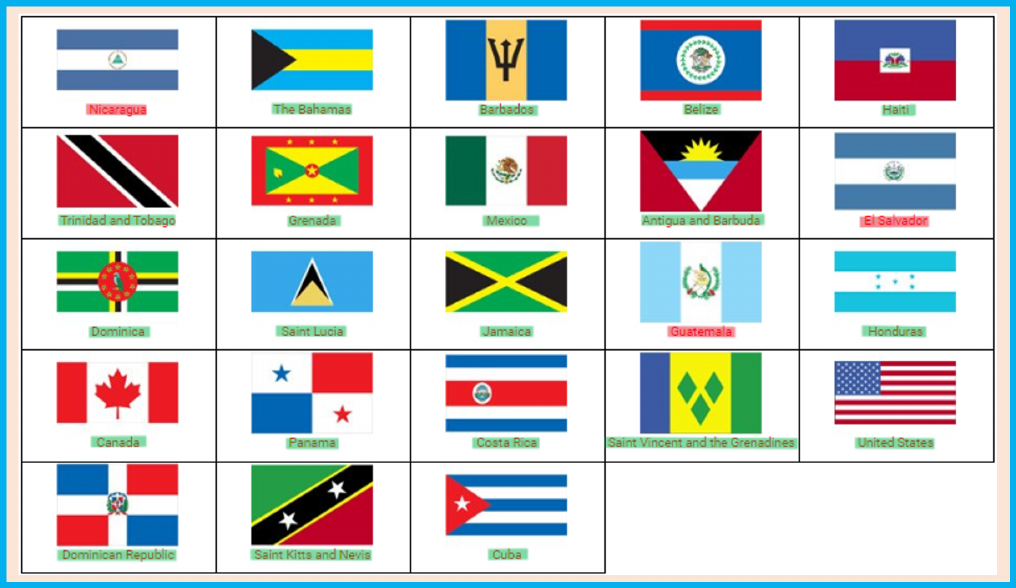}
\caption{Flag identification input \& response.}
\label{fig:prompt_image}
\end{subfigure}
\begin{subfigure}[b]{0.38\textwidth}
\includegraphics[width=\textwidth]{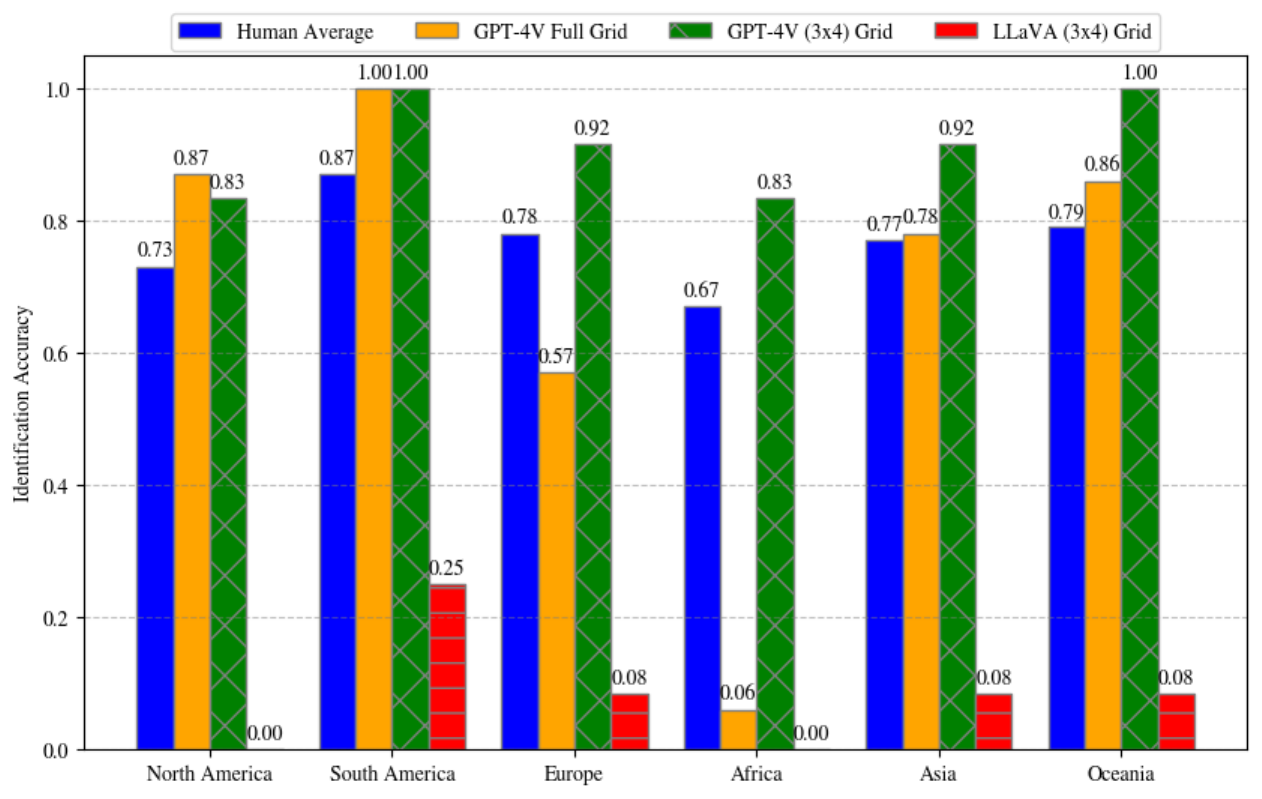}
\caption{Model comparison on flag identification accuracy.}
\label{fig:model_comparison}
\end{subfigure}
\vspace{-0.25cm}
\caption{\textbf{Flag identification}. (a) Presents the structured input utilised for flag identification in the full grid case, along with GPT-4V's responses, where the country names beneath the flags are added after and color-coded to indicate \textcolor{darkgreen}{correct} and \textcolor{red}{incorrect} identifications. (b) Depicts the accuracy in flag identification by different models, with human averages as a reference, noting that human performance data was sourced from Sporcle and may not reflect precise accuracy. }
\label{fig:flags_overview}
\end{figure*}

{
\setlength{\parskip}{-2pt}

The identification of multiple flags in images, a task balancing both the visual and knowledge-based aspects of MLLMs, presents a unique challenge. This is particularly relevant to understanding the geographic biases inherent in these models and their capacity to handle different scales of visual data. Our experiments employed flag images arranged in grid formations sourced from Sporcle quizzes.%

\footnote{\url{https://www.sporcle.com/}}
These flags were divided by continent, cropped, and presented in grids of varying dimensions, challenging the models' scale adaptability and geographic knowledge. The models were engaged with a structured prompt, seen in Fig.~\ref{fig:prompt_image}, intended to simulate a game scenario focused on flag identification:
Notably, in Fig.~\ref{fig:model_comparison}, GPT-4V's proficiency is apparent in its consistent identification accuracy across the standardized 3 x 4 grids, whereas other models falter, often giving non-sensical answers to larger grid formats.

The analysis of the comparative performance data, illustrated in Fig.~\ref{fig:model_comparison}, reveals a discernible geographic bias in model performance, with GPT4-V notably excelling in North and South America. However, its performance on African flags shows a pronounced -61.0\% accuracy differential compared to human benchmarks. While this may suggest a potential underrepresentation of African nations in the training data, it is crucial to consider that other factors may contribute to this outcome, such as grid size, image resolution, and quality.

\subsection{Failure Cases}
Our analysis focuses on tasks where there is sufficient capability in the models to attain reasonable performance. However, we encountered many, at times surprising, failure cases. Settings in which GPT-4V struggled include routing/navigation using maps, drawing and improving country outlines, annotating missing labels on travel maps, estimating population growth from satellite time series, and determining the elevation profile of mountains. Additional details and examples of these cases are included in the \hyperref[app]{Appendix}.

\paragraph{Multiple States.} Using the Cartopy library, we create a series of maps of Africa with 5 countries randomly shaded in different colours (see Fig \ref{fig:shaded_africa}). This task represents a variation of the \textit{State name from outline} experiment, in which additional context is given but there is the added complexity of identifying multiple regions from a single image. We find this task proves challenging: GPT-4V scores $\sim$35\% while LLaVA-1.5 and IDEFICS correctly identify only 5 and 2 African states, respectively. The models are completely unable to correctly identify states when the map is rotated by 180\textdegree.

}

\section{Conclusions}
\label{sec:conclusion}

{

\setlength{\parskip}{-3pt}

We evaluate a selection of state-of-the-art MLLMs on a suite of experiments that explore various vision capabilities in the geographic and geospatial domains. By creating a large proportion of our test data we reduce test contamination, avoiding memorisation. We distill our analysis into key takeaways to clearly inform the research community of MLLM capabilities in this domain. We find that the current generation of MLLMs possesses remarkable abilities to interpret geographic visual data, though struggle with a number of tasks, such as route planning with maps and abstract navigation. Overall, we find GPT-4V to outperform the other models in most settings, and its strong instruction-following ability enables it to attempt a much broader range of tasks. We demonstrate that GPT-4V is able to extract fine-grained details from imagery but is weak at localising and struggles to draw accurate bounding boxes. Of the open-source MLLMs we evaluate, LLaVA-1.5 and Qwen-VL prove to be the most capable and are better suited to some tasks -- \eg object localisation -- than GPT-4V. We additionally introduce two segmentation prompting strategies and release our experimental data as a small-scale geographic benchmark for future evaluation.
\paragraph{Limitations.}
Where possible we seek to maximise the robustness of our experimentation with repeats, consistent prompting and model hyperparameters, normalising inputs and comparisons to ground truth data. However, this robustness is inherently limited by the accessibility of models, specifically API access limits that prevent large-scale quantitative experimentation.

\paragraph{Broader Impacts.}
We demonstrate the strong capabilities of MLLMs to detect fine-grained details from imagery and perform nuanced reasoning, with potential applications in environmental research and disaster response. On the other hand, we highlight numerous cases where the current models are severely lacking, especially in map interpretation. Our analysis suggests there are geographical biases in the data, with weaker performance consistently shown for regions such as Africa that were perhaps less represented in the training distribution.

}

\section*{Acknowledgements}
This work was supported by the UKRI Centre for Doctoral Training in Application of Artificial Intelligence to the study of Environmental Risks (reference EP/S022961/1), an Isaac Newton Trust grant, an EPSRC HPC grant, the Deutsche Forschungsgemeinschaft (DFG, German Research Foundation) - Project-ID 494541002, the Hong Kong Research Grant Council - Early Career Scheme (Grant No. 27208022) and HKU Seed Fund for Basic Research. Samuel would like to acknowledge the support of Z. Novak and N. Novak in enabling his contribution.

{
    \small
    \bibliographystyle{ieeenat_fullname}
    \bibliography{main}
}

\clearpage
\section*{Appendix}
\label{app}
\renewcommand{\thesection}{\Alph{section}}
\setcounter{section}{0}

We structure this Appendix to our main paper in two parts: \textbf{(1)} we provide examples of failure cases in which the models, including GPT-4V, were unable to adequately perform a given task, and \textbf{(2)} we outline specific details of the experiments we discuss, including information regarding hyperparameters and prompts.

\section{Failure cases}

\subsection{Identifying multiple states}

Having created the map images with multiple shaded countries in Africa as mentioned in the main paper, we attain the results for the accuracy of identifying multiple states simultaneously based on the following experimental procedure. We pass each map to the models (GPT-4V, LLaVA-1.5, IDEFICS-80b-Instruct) with the following prompt:

\begin{formattedquote}
    Using the map given above, answer the following:
    Which country is shaded in blue?
    Which country is shaded in orange?
    Which country is shaded in green?
    Which country is shaded in red?
    Which country is shaded in purple?
\end{formattedquote}

We query GPT-4V using the ChatGPT interface and run LLaVA-1.5 and IDEFICS using top\_p set to 0.7 and temperatures of 0.0 and 0.01, respectively. The results we report in the main paper (GPT-4V scoring 19/54, LLaVA-1.5 scoring 5/54 and IDEFICS scoring 2/54) indicate that the task is challenging. The weaker models that score close to 0 are unable to interpret and localise the countries in the image correctly, though in their response descriptions and predictions did mostly correctly identify the continent as Africa. GPT-4V scored well above the chance score but favoured larger states in its predictions and correct answers and rarely correctly identified small states, especially in regions with multiple small states.

\subsection{Map routing}

We explored a variety of different experimental variations involving interpreting journeys using maps. 

In the first setting, we provided GPT-4V with a street map annotated with a start and end marker and asked for a description of a route between the two using the following prompt:

\begin{formattedquote}
    Describe the route to walk from the location marked with the \{start\_marker\} to the location marked with the \{end\_marker\}.
\end{formattedquote}

\begin{figure}
\centering
    \centering
    \includegraphics[height=5cm]{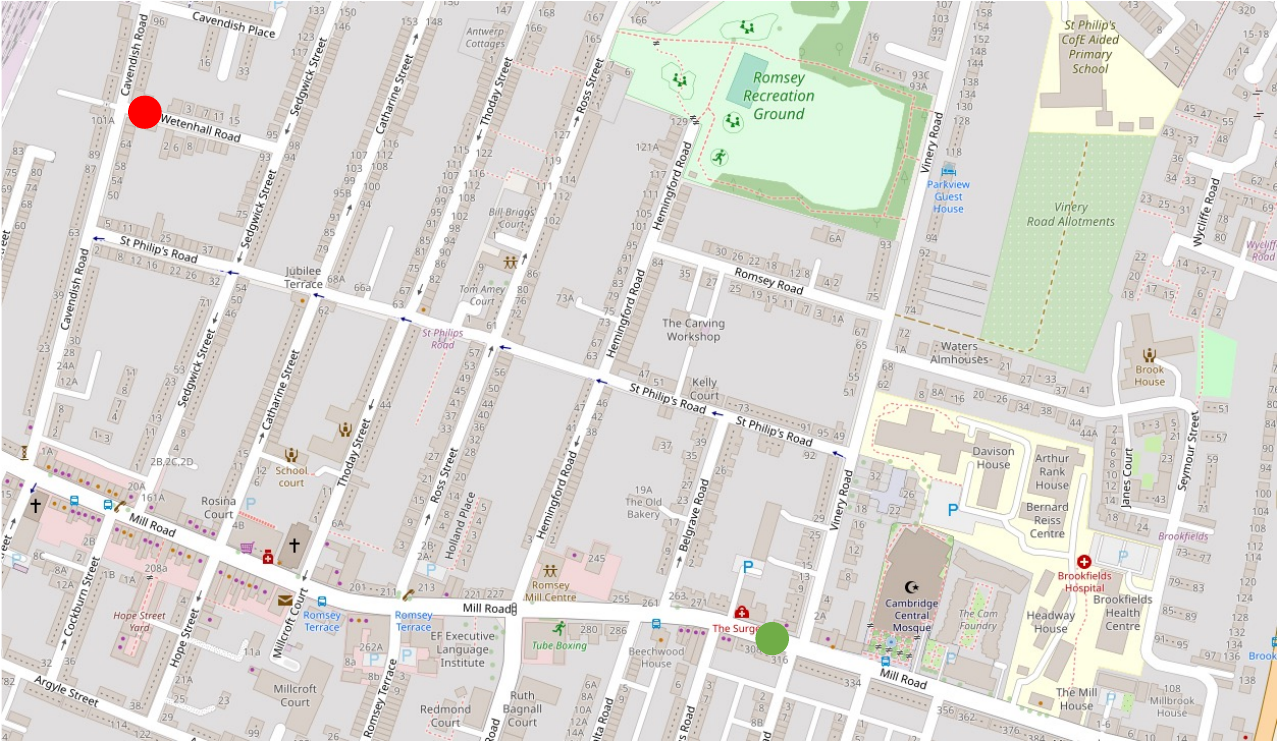}
    \caption{\textbf{Example annotated map input.} An example input OSM map of Cambridge annotated with a journey start (red circle) and end point (green circle).}
    \label{fig:osm}
\end{figure}

However, we found that in every instance ($\sim$25 samples) this resulted in a route that was incorrect and involved hallucinations, such as this response to the annotated OpenStreetMap (OSM)\footnote{\url{https://www.openstreetmap.org/}} map shown in Fig. \ref{fig:osm}:

\begin{formattedresponse}
\begin{enumerate}
    \item Start at the red circle located on Mawson Road.
    \item Head northeast on Mawson Road toward St. Philip's Road.
    \item Turn right onto St. Philip's Road.
    \item Continue on St. Philip's Road until you reach Mill Road.
    \item Cross Mill Road and continue straight onto Romsey Road.
    \item Follow Romsey Road as it curves to the left.
    \item Continue on Romsey Road until you reach the green circle near the junction with Vinery Road.
\end{enumerate}
\end{formattedresponse}

We repeated these experiments across different locations, marker colours and shapes, and map styles, however, incorrect routes were always suggested. 

Another setting that we tried involved providing a series of street-level images along a route followed by a map with a start point and start direction annotated. We then prompted GPT-4V to describe where on the map the final destination was. As in the first setting, the final positions described were inaccurate. 

Finally, we investigated a simple setting consisting of a grid with one cell marked as the start and another as the finish. After passing a description of the setup and instructions to navigate the grid, as well as some examples, we prompted GPT-4V to describe a journey from the start to finish points. Again, we found that GPT-4V was unable to perform this task correctly.

\subsection{Outlines}

\begin{figure*}[ht]
  \centering
  \begin{subfigure}{.22\textwidth}
  \centering
    \includegraphics[height=3cm]{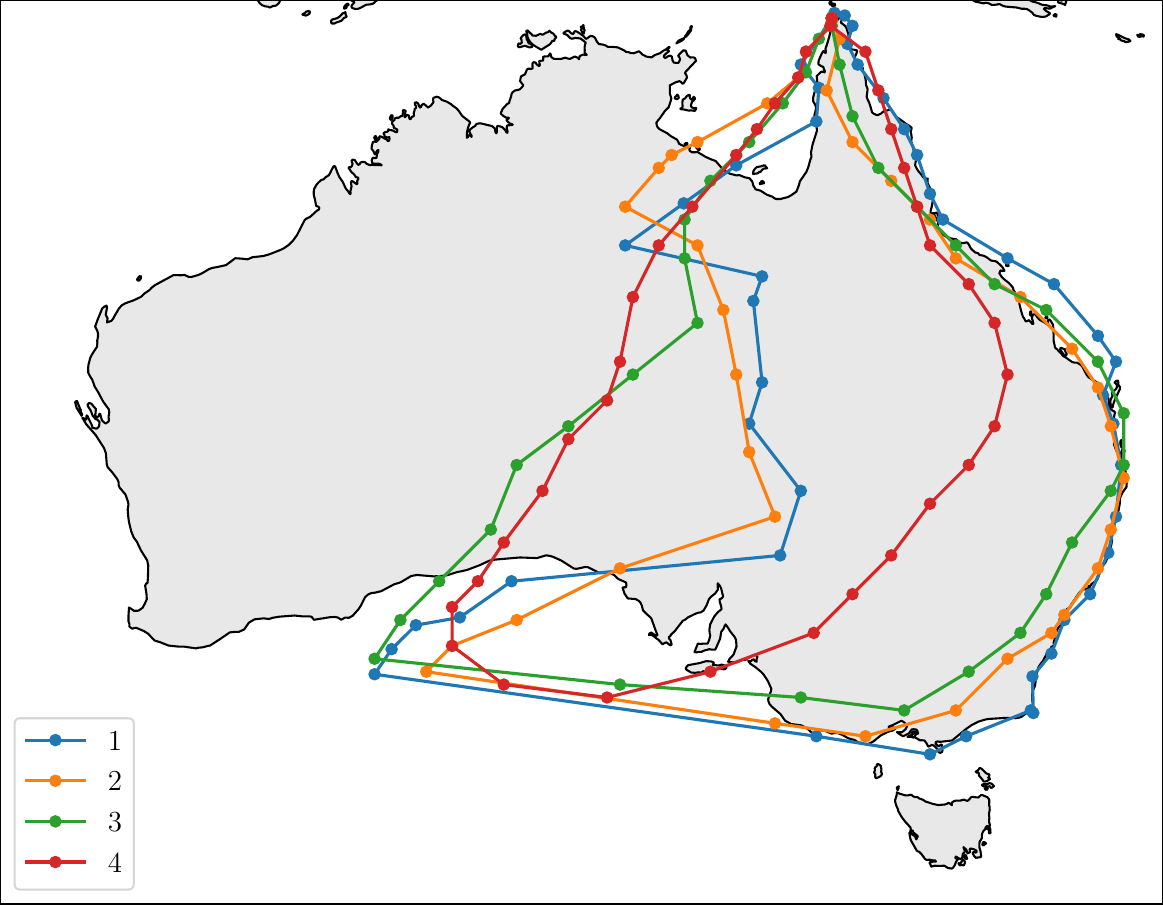}
    \caption{Australia.}
    \label{fig:outline-aus} %
  \end{subfigure}
  \hfill
 \begin{subfigure}{.42\textwidth}
      \centering
  \includegraphics[height=3cm]{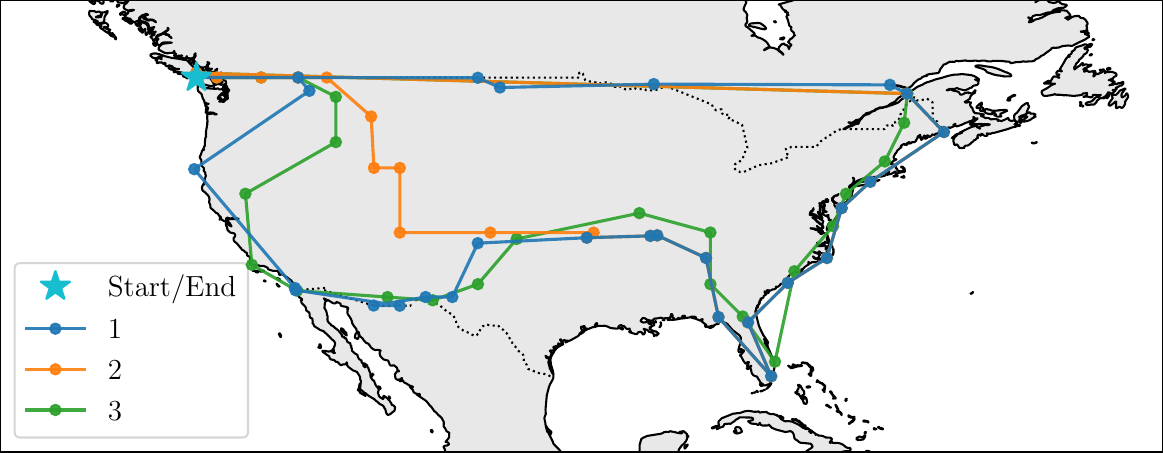}
  \caption{USA.}
  \label{fig:outline-usa}
  \end{subfigure}
    \hfill
 \begin{subfigure}{.35\textwidth}
      \centering
  \includegraphics[height=3cm]{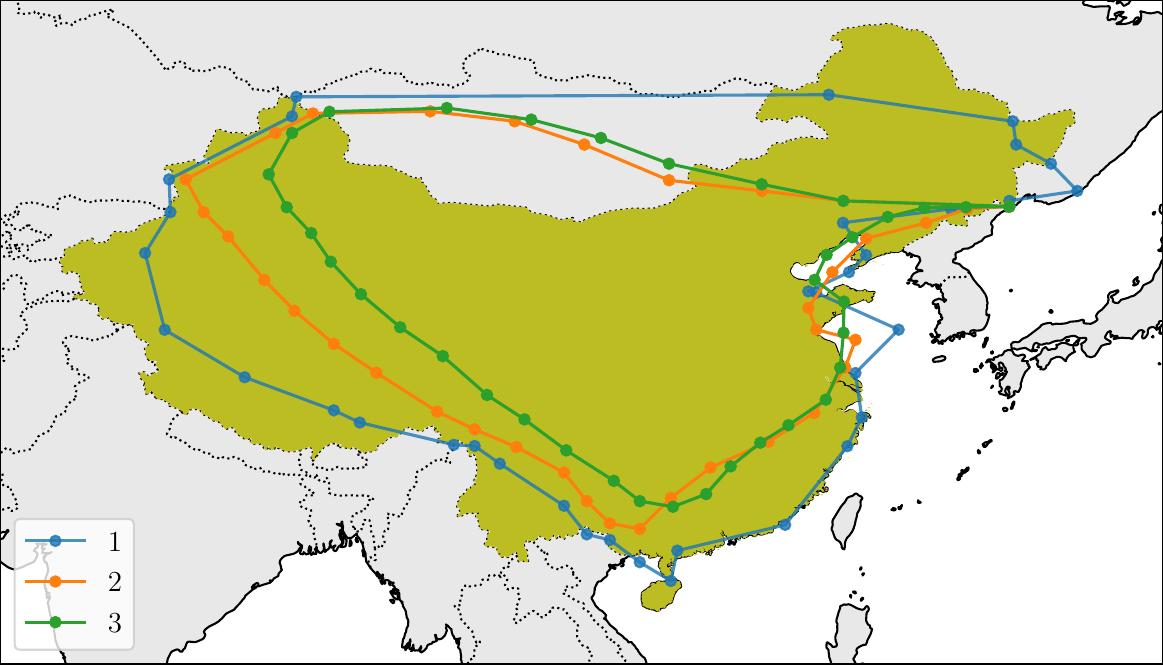}
  \caption{China (shaded).}
  \label{fig:outline-china}
  \end{subfigure}
  \caption{\textbf{Country outlines.} We initially query GPT-4V to provide coordinates for the outlines of specified countries. Then we begin an iterative process of prompting GPT-4V with the provided outlines overlaid on the map. We experiment with different visual markers such as specifying the start and finish point (b) or shading the country boundary (c).}
  \label{fig:outlines}
  
\end{figure*}

In \cite{roberts2023gpt4geo}, qualitative experiments showed that GPT-4 could be prompted to provide reasonably accurate coordinates for the outlines of countries, rivers, etc. For Australia, the authors demonstrated an iterative improvement method in which an initial set of coordinates for the outline of Australia were generated from GPT-4. Then, text feedback was provided to GPT-4 in an attempt to prompt it to provide an improved and more accurate outline. We take inspiration from this experiment however, rather than providing text feedback we plot the provided outlines on a map and pass it to GPT-4V.

We initially query GPT-4V to provide coordinates for the outlines of specified countries, using this text-only prompt:

\begin{formattedquote}
    Provide the latitude and longitude coordinates for an outline of \{country\_name\}. Use approximately 50 points, ensuring that the outline does not overlap itself. Provide the coordinates as a python list in the following format: coordinates = [(LAT\_COORD1, LON\_COORD1), (LAT\_COORD2, LON\_COORD2),...]
\end{formattedquote}

Then we begin an iterative process of providing GPT-4V with the provided outlines overlaid on the map image along with this prompt:

\begin{formattedquote}
    I have plotted this outline (see \{line\_colour\} line). Provide an updated version of the coordinates given the errors in the \{line\_colour\} outline.
\end{formattedquote}

Example country outlines are shown in Fig. \ref{fig:outlines}, illustrating the results of this process. We find that GPT-4V struggles with this task and is unable to consistently improve country outlines. Generally, subsequent sets of coordinate outlines are \textit{less} accurate. We briefly experiment with additional visual markers to aid performance, however, we do not see a noticeable improvement.

\subsection{Networks}

\begin{figure}
    \centering
    \includegraphics[height=7cm]{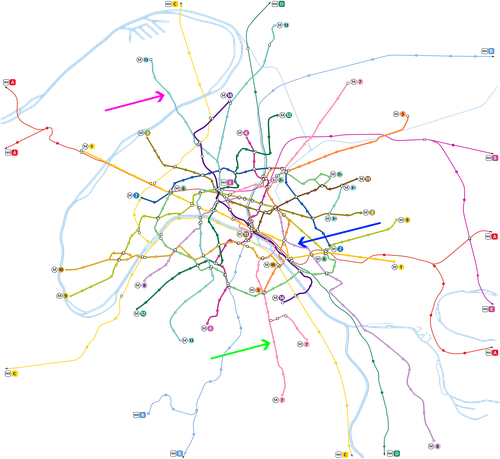}
    \caption{\textbf{Network map labelling}. GPT-4V input: Paris transport network. A high resolution image was provided to GPT-4V.}
    \label{fig:paris-network-map}
\end{figure}

We investigate the ability of GPT-4V to correctly annotate travel network maps. We modify existing network maps by adding visual prompts such as arrows (Fig. \ref{fig:paris-network-map})\footnote{\url{https://commons.wikimedia.org/wiki/File:Paris_Metro_map.svg}}. Next, we pass these images to GPT-4V along with a prompt such as:

\begin{formattedquote}
    The image represents a transport network map for Paris. 3 arrows (purple, blue and green) have been added to the map pointing at specific stations represented by dots. Which station is each arrow pointing to?
\end{formattedquote}

After trying a number of different network maps with different visual prompts and settings (\eg, all but the target stations annotated) this task proved beyond the reaches of GPT-4V. Generally, the model would be unable to provide station predictions, and when it did the answers were consistently incorrect. Given the scale of travel networks, accurate interpretation requires being able to resolve small details.

\subsection{Population estimation}

\begin{figure*}[ht]
\centering
\begin{subfigure}{.48\textwidth}
    \centering
    \includegraphics[width=\textwidth]{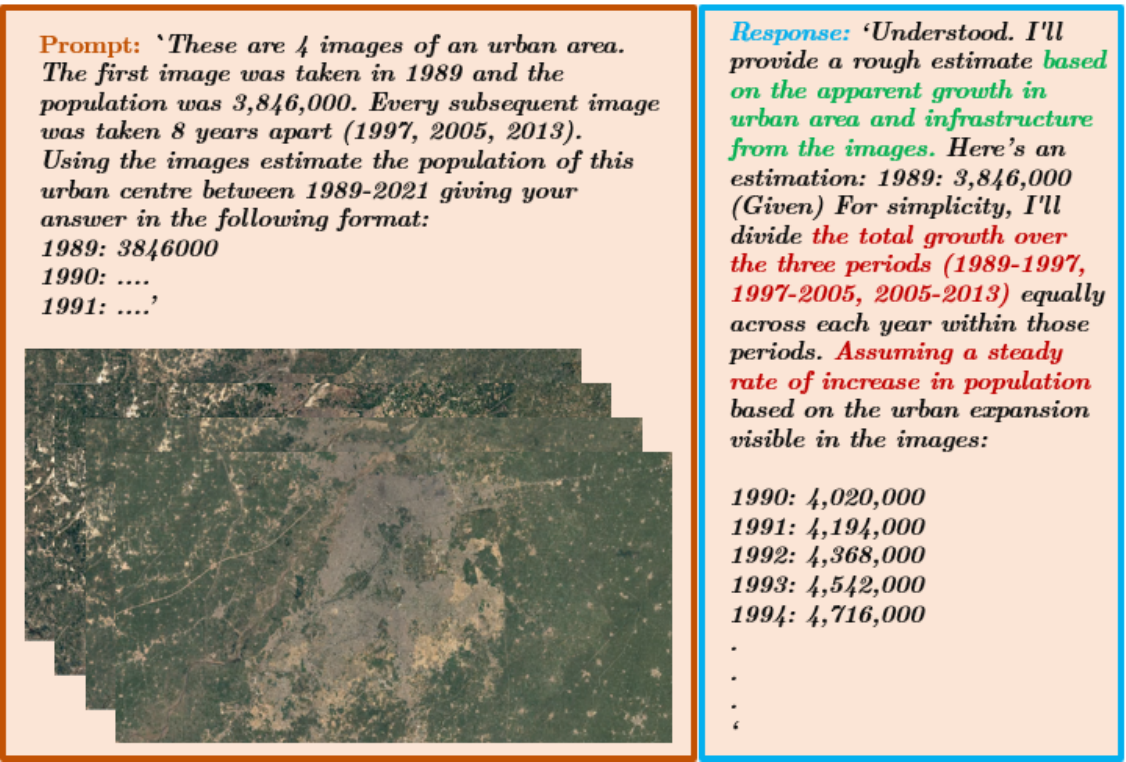}
    \caption{Prompt and response showcasing the model's visual data interpretation ability.}
    \label{fig:population_task_motivation}
\end{subfigure}
\hfill
\begin{subfigure}{.48\textwidth}
    \centering
    \includegraphics[width=\textwidth]{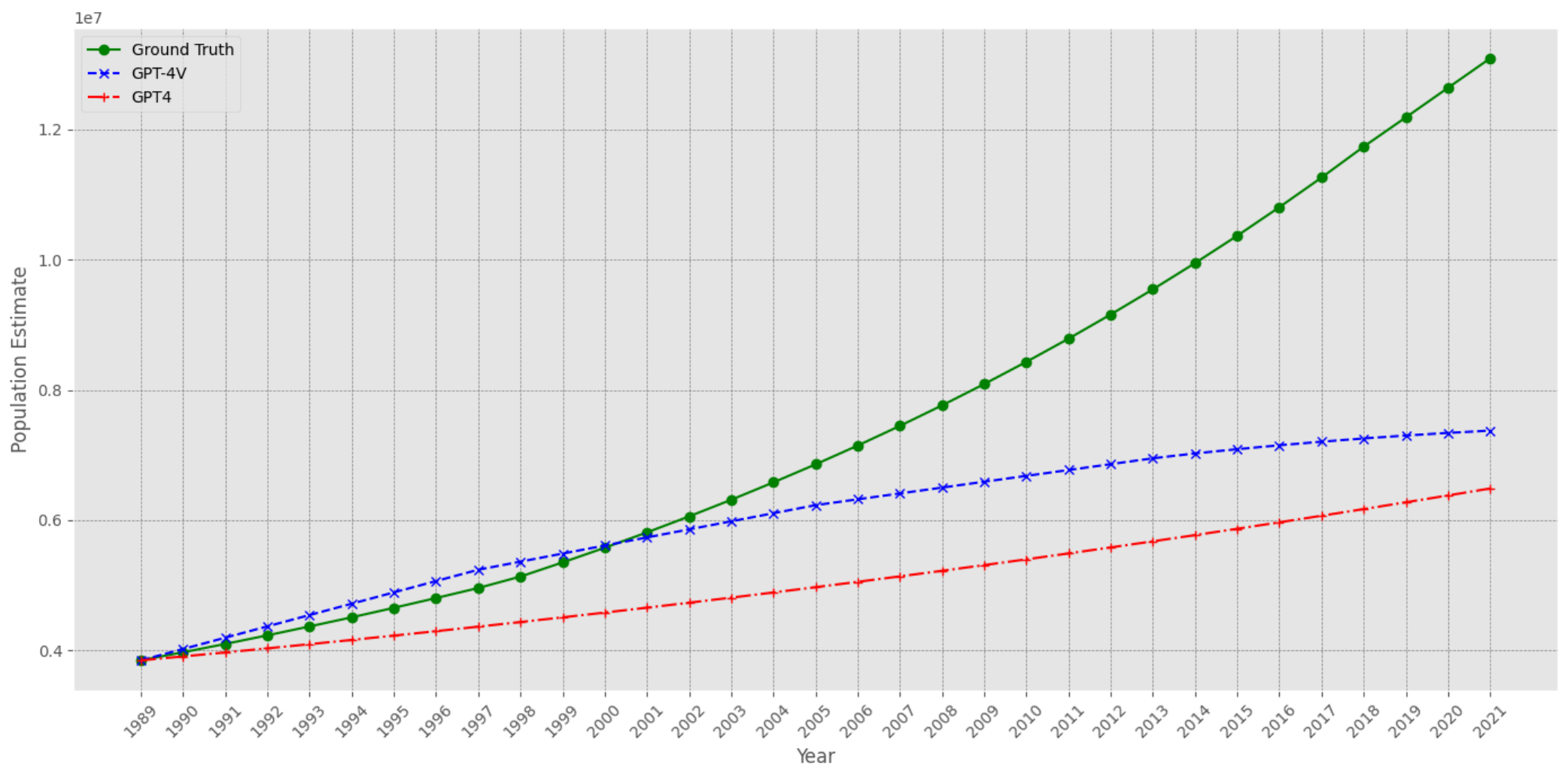}
    \caption{Comparison of population estimates from the GPT-4 and GPT-4V models with the ground truth.}
    \label{fig:population_task_results}
\end{subfigure}    
\caption{\textbf{Population estimation} from satellite imagery task results showing the prompt and model's response (a), and the comparison of the model's estimates to the ground truth (b).}
\end{figure*}

We conducted a population estimation experiment that aimed to assess GPT-4V's ability to predict population growth from satellite imagery. GPT-4V was presented with Google Earth Timelapse\footnote{\url{https://earthengine.google.com/timelapse/}} images of Lahore from 1989 to 2021 and tasked to estimate population changes without external data. The initial population was given, and GPT-4V was prompted to estimate subsequent years' populations based on visual growth cues, as depicted in Fig.~\ref{fig:population_task_motivation}.

The experiment did not succeed as intended. The GPT-4V's population estimates, even with visual data, fell significantly short of the actual figures, as shown in Fig.~\ref{fig:population_task_results}. Although the model with vision performed better than the base model (GPT-4 with just text prompt), it still could not approach the ground truth. This highlights the complexity of the task, which would be challenging even for expert human analysts.

\subsection{Elevation}
\begin{figure}
    \centering
    \includegraphics[width=6cm]{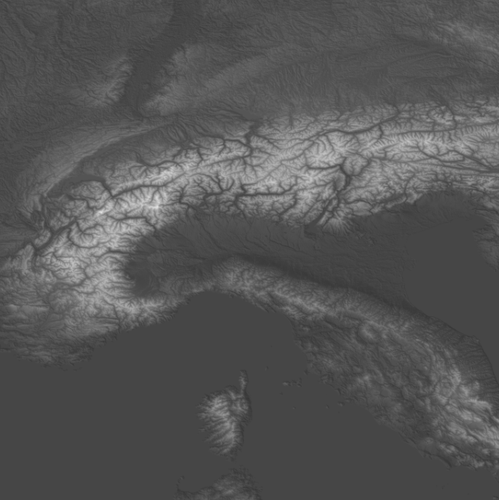}
    \caption{\textbf{Elevation}---Height map of the Alps in Europe.}
    \label{fig:alps_height}
\end{figure}

We asked GPT-4V which mountain range is shown in the image in Fig.~\ref{fig:alps_height} through the prompt:
\begin{formattedquote}
Which mountains are depicted in this height map?
\end{formattedquote} 
Despite first refusing to answer, after being explicitly asked to make a guess GPT-4V mentioned the following (the actual answer is longer)
\textit{The Appalachian Mountains in North America, the Alps in Europe and the Andes in South America}.
We consider this a failure case as the model mentioned the correct answer (Alps in Europe) as an option among many incorrect answers.

\section{Experimental details}

Aside from cases where we interact with GPT-4V via the ChatGPT interface (and hence have no control over the model hyperparameters), we set the value of top\_p to 0.7 (0.8 for Qwen-VL) for all experiments and models. We use different values for the temperature parameter depending on the experiment, the specific values used for each are detailed below.

\subsection{Localisation}

\paragraph{Preprocessing.}
We take the GeoLocation dataset \cite{kaggle_geoguessr} and randomly sample one image per sovereign state (other states are dropped). Each image is preprocessed to remove potentially confusing watermarks (\eg, overlaid map). Rather than completely removing all watermarks, which will significantly decrease the image size and field of view, we instead take a crop removing the rightmost 284 pixels.

\paragraph{Prompts.}
For the localisation (GeoGuessr) experiment, we pass the models a single image along with the following prompts. Where relevant, model temperature is included in parenthesis.

\noindent\underline{GPT-4V}
\begin{formattedquote}
    Which country is this picture taken in? If you are not certain then provide an educated guess of a specific country. Let's think step by step.
\end{formattedquote} 

\noindent\underline{LLaVA-1.5} (\textit{temperature}=0.001)
\begin{formattedquote}
    Which country is this picture taken in? If you are not certain then provide an educated guess of a specific country. Let's think step by step. You must provide a predicted country.
\end{formattedquote}

\noindent\underline{Qwen-VL} (\textit{temperature}=0.1)
\begin{formattedquote}
    Question: Which country is this picture taken in? If you are not certain then provide an educated guess of a specific country. Let's think step by step. You MUST provide a single predicted country. Answer:
\end{formattedquote}

followed by (if an answer is not provided)

\begin{formattedquote}
    Pick a single most likely country. Make an educated guess if you are unsure.
\end{formattedquote}

then

\begin{formattedquote}
    Pick a single most likely country.
\end{formattedquote}

then 

\begin{formattedquote}
    Pick the most likely continent.
\end{formattedquote}

\noindent\underline{InstructBLIP} (\textit{temperature}=0.1)
\begin{formattedquote}
    Which country is this picture taken in? If you are not certain then provide an educated guess of a specific country. Let's think step by step. You must provide a predicted country.
\end{formattedquote}

\subsection{Remote sensing -- classification}

We randomly select the following two datasets from each task in the SATIN metadataset \cite{roberts2023satin}: NASC-TG2, SAT-6 (Task 1), WHU-RS19, EuroSAT (Task 2), RSI-CB256, Million-AID (Task 3), MultiScene,	AID-MultiLabel (Task 4), Post Hurricane, UTSC\_SmokeRS (Task 5), and Brazilian Coffee Scenes, Brazilian Cerrado-Savanna Scenes (Task 6). For each task, we take a class-balanced subset of 12 images. Images were passed individually to each model with the exception of GPT-4V, in which images were passed in batches of 4 via the ChatGPT interface.

The following prompts were used (where \{class\_labels\} is a comma-separated sequence of the target classes for the dataset):

\noindent\noindent\underline{GPT-4V}: Tasks 1-3,5-6
\begin{formattedquote}
Classify each of these images with a single label from this set: \{class\_labels\}. You must provide a predicted label for each image.
\end{formattedquote}
for Task 4
\begin{formattedquote}
Classify each of these images with labels from this set: \{class\_labels\}.
\end{formattedquote}

\noindent\underline{LLaVA-1.5} (\textit{temperature}=0.7): Tasks 1-3,5-6
\begin{formattedquote}
    Classify the image with a single label from this set: \{class\_labels\}. You must provide a single label.
\end{formattedquote}

for Task 4
\begin{formattedquote}
    Classify the image with labels from this set: \{class\_labels\}. You must only use labels from this set.
\end{formattedquote}

\noindent\underline{Qwen-VL} (\textit{temperature}=0.75): Tasks 1-3,5-6
\begin{formattedquote}
    Question: Classify the image with a single label from this set: \{class\_labels\}. You must provide a single label. If you are unsure make an educated guess. Label:
\end{formattedquote}

for Task 4
\begin{formattedquote}
    Question: Classify the image with labels from this set: \{class\_labels\}. You must provide predicted labels for the image. Answer:
\end{formattedquote}

\subsection{Remote sensing -- change detection}
The 4 time series images shown in the main paper were passed to GPT-4V in a single batch along with the prompt shown in the main paper.

\subsection{Remote sensing -- segmentation}
The prompts used for the segmentation example are as follows, using the target land cover classes from the LoveDA dataset \cite{wang2022loveda}.

Grid segementation:
\begin{formattedquote}
Segment the image into the following 7 land cover classes: background (1), building (2), road (3), water (4), barren (5), forest (6), agriculture (7). Display the results as a 15x15 table with each cell labelled with one of the 7 class labels. Don't include any column or row labels.
\end{formattedquote}

SVG segmentation:
\begin{formattedquote}
Segment the image into the following 7 land cover classes: background (1), building (2), road (3), water (4), barren (5), forest (6), agriculture (7). Provide the code for an SVG that displays the segmentation map.
\end{formattedquote}

\subsection{Remote sensing -- bounding boxes}

We derived bounding box coordinates using the following prompts:

\noindent \underline{GPT-4V}
\begin{formattedquote}
    Please follow the instructions 1. Tell me the size of the input image; 2. Localize each \{object\_class\} in the image using a bounding box.
\end{formattedquote}

\noindent \underline{LLaVA-1.5}, \underline{IDEFICS}, \underline{Qwen}
\begin{formattedquote}
    Localize the \{object\_class\} in the image using a bounding box. 
\end{formattedquote}

\noindent \underline{InstructBLIP}
\begin{formattedquote}
    Localize the \{object\_class\} in the image using a bounding box. Bounding box coordinates: 
\end{formattedquote}

\noindent \underline{Kosmos-2}
\begin{formattedquote}
    <grounding><phrase> \{object\_class\} </phrase>
\end{formattedquote}

\subsection{Remote sensing -- counting}

We used the same prompt given in the main paper for each model.

\subsection{Mapping -- region identification}

\paragraph{City Maps}
We based our analysis on the following 16 cities:\\ \textit{Madrid, Naples, Cairo, Lagos, Buenos Aires, Rio de Janeiro, Mexico City,
New York City, San Francisco, Shanghai, Taipei, Mumbai, Tokyo, Stockholm, Cape Town, Vladivostok}.\\

For GPT-4V, we sequentially fed single images into a conversation, which we started by this prompt:
\begin{formattedquote}
    I'll show you the map of a city and you tell me its name. Make only a single guess. The top of the image is facing north.
\end{formattedquote}

For the other models we independently asked about single examples using the same prompt (for LLaVA-1.5) or the following slightly amended prompt for Qwen and InstructBLIP:
\begin{formattedquote}
    Question: Guess the city that is shown in the map. Make only a single guess. The top of the image is facing north. City: 
\end{formattedquote}

\paragraph{Islands and Water Bodies}

We based our analysis on the following 16 water bodies and islands:\\ \textit{Strait of Gibraltar, Balearic Islands, Sicily, Strait of Hormuz, Bahamas, Galapagos, Samborombón Bay, Falkland Islands, South Georgia, Tasmania, Hawaiian Islands, \"{O}resund, Strait of Malacca, 
Tahiti, Spitsbergen, Bering Strait}.\\
The selection of places and the exact map crop (including zoom level) were carried out manually. 

For GPT-4V, we sequentially fed single images into a conversation, which we started by this prompt:
\begin{formattedquote}
    I'll show you the shape of an island or water body and you tell me its name. Make only a single guess. The top of the image is facing north.
\end{formattedquote}
For the other models we independently asked about single examples using the same prompt (for LLaVA-1.5) or the following slightly amended prompt for Qwen and InstructBLIP:
\begin{formattedquote}
    Question: Guess the island or water body shown in the image. Make only a single guess. The top of the image is facing north. Island/Water body: 
\end{formattedquote}

\paragraph{Country Shapes}
We used the following 16 states for this experiment:\\
\textit{Indonesia, Jordan, Vietnam, Poland, Russia, Belgium, Haiti, Mexico, United States of America, Ethiopia, Tanzania, Malawi, Australia, Colombia, Venezuela, Peru}.\\
They were selected randomly from all countries with a population above 10M while balancing over continents (3 states per continent, except oceania with 1).

For GPT-4V, batches of four images were provided at the same time together with this prompt:
\begin{formattedquote}
    Guess the country by the provided shape. Output nothing but a single guess for each shape.
\end{formattedquote}
For the other models we independently asked about single examples using the same prompt (for LLaVA-1.5) or the following slightly amended prompt for Qwen and InstructBLIP:
\begin{formattedquote}
    Question: Guess the country by the provided shape. Output nothing but a single guess for each shape. Country: 
\end{formattedquote}

\subsection{Mapping -- Localisation: map $\rightarrow$ real-world}

The continent maps were created by taking a crop of 50 degrees latitude and longitude extent. (Note, in Fig. \ref{fig:projections_output} the map showing the predicted positions of a set of test points in Africa has been enlarged beyond the extend of the input image in order to include the predicted positions.) To position the coloured test points we employed stratified random sampling within the map extent with a margin of 2 degrees from the edges (to avoid overlap with the map edge). For a given projection, we queried GPT-4V via the ChatGPT interface in a single chat window using the following prompt on two batches of 3 images (each representing a different continent). Batch one included Africa, Asia and Europe, while Batch two included North America, Oceania and South America.  

\begin{formattedquote}
    For each of the images, please estimate the Latitude/Longitude positions of the coloured points on the map to 2 decimal places. A point is included on the map for each of the following colours so provide a position estimate for each. Provide the coordinates as a list in the following format:
    estimated\_coordinates = [
    [Longitude\_Blue, Latitude\_Blue], \# Blue point
    [Longitude\_Orange, Latitude\_Orange], \# Orange point
    [Longitude\_Green, Latitude\_Green], \# Green point
    [Longitude\_Red, Latitude\_Red], \# Red point
    [Longitude\_Purple, Latitude\_Purple], \# Purple point
    [Longitude\_Brown, Latitude\_Brown], \# Brown point
    [Longitude\_Pink, Latitude\_Pink], \# Pink point
    [Longitude\_Gray, Latitude\_Gray], \# Gray point
    [Longitude\_Olive, Latitude\_Olive], \# Olive point
    [Longitude\_Cyan, Latitude\_Cyan], \# Cyan point
    ]
\end{formattedquote}

Occasionally, follow-up prompts were needed to ensure a position prediction was given for every coloured point.

\subsection{Mapping -- Localisation: real-world $\rightarrow$ map}

In addition to the results reported in the main paper, we experimented with different grid sizes, grid line colours and map projections but were unable to find a configuration that improved the overall accuracy significantly. Other models showed an inability to correctly interpret the task and provide valid grid coordinates.

\subsection{Flags}
\label{subsec:refined_flags}

\begin{figure*}[ht]
\centering
\includegraphics[width=\textwidth]{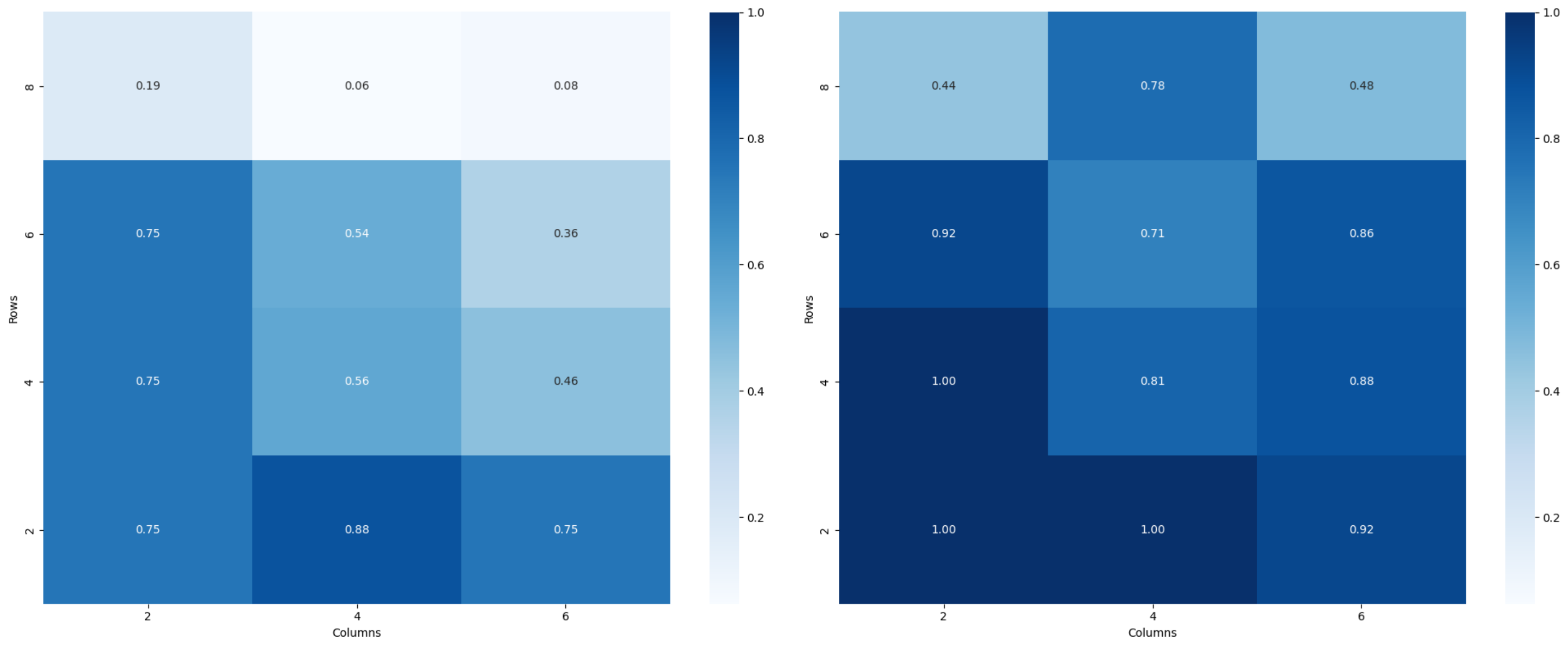}
\caption{\textbf{Flag Identification Performance}. Heatmaps detail GPT-4V's accuracy in identifying flags from cropped grid segments of Africa \textbf{[L]} and Asia \textbf{[R]}. Darker cells indicate higher identification accuracy.}
\label{fig:cropped_flags_heatmap}
\end{figure*}

We examined GPT-4V's capacity for flag identification by presenting cropped segments from the original (from Sporcle) full grids of African and Asian flags. The crops correspond to the specific row and column layout indicated in the heatmaps (Fig.~\ref{fig:cropped_flags_heatmap}), which reflect the model's accuracy scores for each segment.

The models were provided with the following prompt with corresponding grid size parameters based on the crop size. The prompt directed the models as follows:

\noindent\underline{LLaVA-1.5} (\textit{temperature}=0.2)

\begin{formattedquote}
You will be playing the following game with me: Flags of \{CONTINENT\}. Can you name the flags of \{CONTINENT\}? Respond in the following format. Think step by step during this game. Note that the grid is \{ROWS\} (row) x \{COLUMNS\} (column). Replace xxxx with your guess of the flag. RESPONSE (x row, y column): (1, 1) - xxxx, (1, 2) – xxxx
\end{formattedquote}
This focused approach aimed to discern GPT-4V's recognition efficacy when confronted with a condensed subset of visual information. Preliminary results indicate variability in the model's performance across different segments, with some showing high accuracy while others underperform. The heatmaps encapsulate this performance, highlighting areas of both strength and potential weakness within the model's identification capabilities.

While these findings might suggest regional biases in flag identification, conclusions about model bias should be drawn with caution. The variability in accuracy may also be influenced by factors such as image processing or the distinctive features of certain flags. As such, these results may warrant further investigation to better understand the underlying causes of performance discrepancies and to determine if they indeed reflect biases within the model's training data or recognition abilities.

\end{document}